\documentclass[10pt,journal,compsoc]{IEEEtran}

\usepackage{graphicx}
\usepackage{subfigure}
\usepackage{enumitem}
\usepackage{amsmath}
\usepackage{amsthm}
\usepackage{amssymb}
\usepackage{bm}
\usepackage{multirow}
\usepackage{url}
\usepackage{color}
\usepackage{nicefrac}
\usepackage{float}
\usepackage{threeparttable}
\usepackage{booktabs}
\usepackage[normalem]{ulem}
\usepackage{bbm}

\newtheorem{definition}{Definition}

\usepackage{array}
\newcolumntype{L}[1]{>{\raggedright\let\newline\\\arraybackslash\hspace{0pt}}m{#1}}
\newcolumntype{C}[1]{>{\centering\let\newline  \\\arraybackslash\hspace{0pt}}m{#1}}
\newcolumntype{R}[1]{>{\raggedleft\let\newline \\\arraybackslash\hspace{0pt}}m{#1}}

\newcommand{\M}[1]{\mathcal{#1}}

\ifCLASSOPTIONcompsoc
  \usepackage[nocompress]{cite}
\else
  \usepackage{cite}
\fi

\ifCLASSINFOpdf
\else
\fi

\usepackage{algorithm}
\usepackage{algorithmicx}
\usepackage{algpseudocode}

\begin{document}

\title{Searching a High Performance Feature Extractor for Text Recognition Network}

\author{Hui Zhang,~\IEEEmembership{}
        Quanming Yao,~\IEEEmembership{Member, IEEE}
        James T. Kwok,~\IEEEmembership{Fellow, IEEE}
        \\
        Xiang Bai,~\IEEEmembership{Senior Member, IEEE }
\IEEEcompsocitemizethanks{\IEEEcompsocthanksitem 
	H. Zhang is with 4Paradigm Inc;
	\IEEEcompsocthanksitem
	Q. Yao is with  Department of Electronic Engineering, Tsinghua University;
	\IEEEcompsocthanksitem
	J. Kwok is with Department of Computer Science and Engineering, 
	Hong Kong University of Science and Technology.
	\IEEEcompsocthanksitem
	X. Bai is with Department of Electronics and Information Engineering, 
	Huazhong University of Science and Technology.
	\IEEEcompsocthanksitem
	Corespondance is to Q.Yao at qyaoaa@tsinghua.edu.cn
}
\thanks{Manuscript received xxx; revised xxx.}}

\markboth{---}{---}

\IEEEtitleabstractindextext{
\begin{abstract}
Feature extractor plays a critical role in text recognition (TR), but customizing its architecture is relatively less explored due to
expensive manual tweaking. In this work, inspired by the success of neural architecture search (NAS), we propose to search for suitable feature extractors. We design a domain-specific search
space by exploring principles for having good feature extractors. The space includes a 3D-structured space for the spatial model and a transformed-based space for the sequential model.  As the space is huge and complexly structured, no existing NAS algorithms can be applied. We propose a two-stage algorithm to effectively search in the space. In the first stage, we cut the space into several blocks and progressively train each block with the help of an auxiliary head. We introduce the latency constrain into the second stage and search sub-network from the trained supernet via natural gradient descent. In experiments, a series of ablation studies are performed to better understand the designed space, search algorithm, and searched architectures. We also compare the proposed method with various state-of-the-art ones on both hand-written and scene TR tasks. Extensive results show that our approach can achieve better recognition performance with less latency.
\end{abstract}

}

\maketitle

\IEEEdisplaynontitleabstractindextext

\IEEEpeerreviewmaketitle


\section{Introduction}
\label{sec:intro}

\IEEEPARstart{T}ext recognition (TR)~\cite{long2018scene,zhu2016scene}, 
which targets at extracting text from document or natural images,
has attracted great interest from both the industry and academia.
TR is a challenging problem~\cite{shi2018aster}
as the
text can have diverse
appearances and
large variations in size, fonts, background, writing style, and layout.

\begin{figure}[t]
	\centering
	\includegraphics[width=0.70\linewidth]{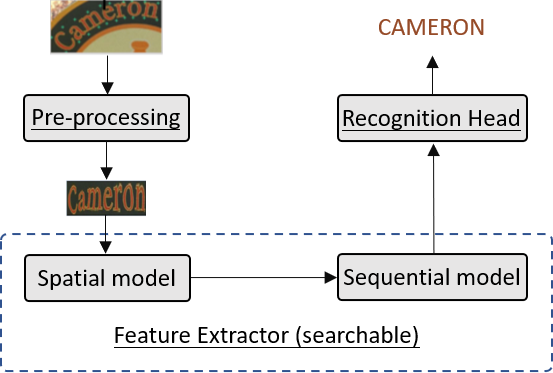}
	\vspace{-10px}
	\caption{The typical text recognition~(TR) pipeline. 
In this paper, we focus on the
search
of a good feature extractor.}
	\vspace{-10px}
	\label{fig:general_structure}
\end{figure}

Figure~\ref{fig:general_structure} shows the typical TR pipeline. It
can be divided into three modules: (i)
An 
optional 
pre-processing module
which transforms the input image to a more recognizable form.
Representative methods include rectification~\cite{shi2018aster}, 
super-resolution~\cite{wang2020scene} and
denoising~\cite{luo2021separating}.
(ii) A feature extractor,
which extracts features from the text images. Most of them 
\cite{shi2018aster,yue2020robustscanner,wan2020textscanner}
use a combination of
convolutional neural networks (CNNs) and 
recurrent neural networks (RNNs).
The CNN extracts spatial features 
from the image, which are then
enhanced
by the RNN 
for the generation of robust sequence features~\cite{shi2016end,baek2019wrong}.
(iii)
A recognition head,
which outputs the character sequence. Popular choices are based on
connectionist temporal classification~\cite{shi2016end}, 
segmentation~\cite{liao2019scene},
sequence-to-sequence attention~\cite{shi2018aster},  and
parallel attention~\cite{yue2020robustscanner}.

\begin{table*}[ht]
	\centering
	\caption{Some well-known hand-designed 
		text recognition (TR) algorithms
		with
		NAS-based methods.}
	\vspace{-10px}
	\begin{tabular}{c|c | c c | c | c || c}
		\toprule
		& \multirow{2}{*}{\textbf{method}} & \multicolumn{2}{c|}{\textbf{spatial model}} & \multirow{2}{*}{\textbf{sequential model}} & \textbf{search}    & \textbf{deployment-} \\
		&                                  & \textbf{downsampling path } & \textbf{conv layer}          &                                            & \textbf{algorithm} & \textbf{aware}       \\ \midrule
		hand-   &      CRNN~\cite{shi2016end}      & fixed             & vgg~\cite{simonyan2014very}                     & BiLSTM                                     & ---                & $\times$             \\
		designed 
		&    ASTER~\cite{shi2018aster}     & fixed             & residual~\cite{he2016deep}                & BiLSTM                                     & ---                & $\times$             \\
		&    GFCN~\cite{Coquenet2020}      & fixed             & gated-block~\cite{Coquenet2020}             & -                                          & ---                & $\times$             \\
		&   SCRN~\cite{yang2019symmetry}   & fixed             & residual~\cite{he2016deep}                & BiLSTM                                     & ---                & $\times$             \\ \midrule
		&  STR-NAS~\cite{hong2020memory}   & fixed             & searched                & BiLSTM                                     & grad.              & $\times$             \\
		NAS     & AutoSTR~\cite{zhang2020autostr}  & two-dim           & searched                & BiLSTM                                     & grid+grad.         & $\times$             \\
		&              TREFE (proposed)              & two-dim           & searched                & searched                                   & NG                 & $\surd$              \\ 
		\bottomrule
	\end{tabular}
	\label{tab:txtreg}
	\vspace{-10px}
\end{table*} 

The feature extractor plays an important role in TR.
For example, 
in~\cite{shi2018aster,baek2019wrong},
significant performance gains 
are observed
by simply replacing the feature extractor 
from VGG
\cite{simonyan2014very} 
to ResNet
\cite{he2016deep}. 
Furthermore, 
the feature extractor often requires a lot of compute and 
storage~\cite{li2019show,yang2019symmetry}. 
However, while a lot of 
improvements
have been proposed 
for the 
pre-processing module and TR head,
design of the feature extractor is less explored.
Existing methods~\cite{shi2018aster,yue2020robustscanner,wan2020textscanner} 
often
directly use CNNs and RNNs that are originally designed for other tasks
(Table~\ref{tab:txtreg}),
without specially tuning them for TR.
Examples include the direct use of ResNet that is originally used for image classification,
and BiLSTM~\cite{hochreiter1997long} from language translation.
Moreover, 
TR systems
often have
inference latency constraints
on deployment to real-world devices
\cite{chen2020fasterseg, guo2020single}.
However,
existing 
designs
do not explicitly take this into account.
Manually tweaking  the TR system 
to
satisfy the latency constraint 
while 
maintaining a high recognition accuracy
can be hard~\cite{guo2020single}.

Recently, it has been shown that
neural architecture search (NAS)~\cite{elsken2019neural}
can produce
good network architectures 
in 
tasks
such as computer vision 
(e.g.,
image classification~\cite{DARTS,pham2018enas},
semantic segmentation~\cite{liu2019auto} 
and object detection~\cite{chen2019detnas}).
Inspired by this, 
rather than relying on experts to design architectures,
we propose 
the use of 
one-shot NAS 
\cite{Cai19Proxylessnas,DARTS,pham2018enas}
to search for  a
high-performance TR feature extractor.
Specifically,
we first design TR-specific search spaces
for the 
spatial and sequential feature
extractors.
For the spatial component, 
the proposed search space allows selection of both the convolution type 
and feature downsampling path.
For the sequential component, 
we propose to use transformer instead,
which
has better parallelism than
the 
BiLSTM
commonly
used in TR.
However, we find the vanilla transform is hard to beat BiLSTM.
Thus,
we further explore the recent advances of the Transformer, 
and search for variants of the transformer~\cite{VaswaniSPUJGKP17}.

As the resultant supernet is huge,
we propose to use the
two-stage 
one-shot NAS approach
\cite{guo2020single,CaiGWZH20}.
In the first stage, 
inspired by the success of progressive layer-wise training of deep networks~\cite{bengio2007greedy,hinton2006fast},
we train the supernet
in a greedy block-wise manner.
In the second stage, 
instead of using evolutionary algorithms or random  search as in~\cite{li2019random,guo2020single,CaiGWZH20},
we use natural gradient descent~\cite{amari1998natural}
to 
more efficiently
search for a compact architecture 
from the trained supernet.
Resource constraints on the deployment environment
can also be easily added
in this stage, 
leading to 
models that are deployment-aware.
Extensive experiments on 
a number of standard benchmark datasets
demonstrate that 
the resultant TR model 
outperforms the state-of-the-arts in terms of both accuracy and
inference speed.

Concurrently,
Hong \textit{et al.}~\cite{hong2020memory} also considered the use of NAS in scene text recognition.
However, they 
only search for the convolution operator,  while
we 
search for both the spatial feature and 
sequential feature extractors with deployment constraints (see Table~\ref{tab:txtreg}).
As a result,
our search space is much larger and more complex,
and a straightforward application of existing NAS algorithms
is not efficient.



This paper is based on an earlier conference version ~\cite{zhang2020autostr},
with the following major extensions:
\begin{itemize}[leftmargin=*]
\item
The search space
is expanded by including search on 
the sequential model (Section~\ref{sec:rnn_search_space}) 
and allow more possibility of downsampling path 
for the spatial model (Section~\ref{sec:path_search_space}).
Experiments in
Section~\ref{sec:exp:comp}
demonstrates that 
both parts 
contribute the performance improvement
over AutoSTR.

\item 
The search algorithm
is redesigned (Section~\ref{sec:search_alg}).
In~\cite{zhang2020autostr},
the search is performed in a step-wise search space that 
cannot explore all candidate architectures 
and assess the real deployment performance.
In this paper,
we first construct a supernet that 
contains all candidates in the search space 
(Section~\ref{sec:design_supernet}),
which also allows 
direct evaluation of the deployment performance
of each compact architecture 
(Section~\ref{ssec:search}).
In order to train the supernet easily, 
we propose a progressive training strategy (Section~\ref{ssec:train_supernet}).
In the search stage for candidate structures, 
we propose to use natural gradient descent~\cite{amari1998natural} and 
introduce latency constraint for deployment awareness (Section~\ref{ssec:search}).

\item
Experiments are much more extensive.
While the conference version
\cite{zhang2020autostr}
only 
evaluates on scene text  datasets, 
in Section~\ref{sec:exp:sota}
we
also evaluate on handwritten text datasets.
Besides,
more recent 
state-of-the-art baselines
are included.
We also provide a detailed examination of the 
searched architectures (Section~\ref{sec:aba1})
and 
search algorithm (Section~\ref{sec:exp:proce}).
\end{itemize}


\section{Related Works}
\label{sec:rel}


\subsection{Text Recognition~(TR)}
\label{sec:rel:tr}

In the last decade,
deep learning has been highly successful  and 
achieves remarkable performance on the recognition  of
handwritten text~\cite{Coquenet2020,yousef2020origaminet} and 
scene text~\cite{shi2018aster,yan2021primitive}.  
However, 
the large variations in size, fonts, background, writing style, and layout
still make TR from images a challenging problem~\cite{ye2014text}.
Existing TR methods usually has three modules:
(i) pre-processing module, 
(ii) feature extractor, and (iii) TR head
(Figure~\ref{fig:general_structure}).


\subsubsection{Pre-Processing Module}

The pre-processing module makes the input text image 
more easily recognizable. 
Shi \textit{et al.}~\cite{shi2016robust,shi2018aster}
uses a learnable 
Spatial Transformer Network (STN)
to 
rectify the irregular text to a canonical form before recognition.
Subsequent methods~\cite{yang2019symmetry,zhan2019esir}
further improve the transformation to achieve
more accurate
rectifications.
Wang \textit{et al.}~\cite{wang2020scene} 
introduces a super-resolution transformation 
to make blurry and low-resolution images clearer.
Luo \textit{et al.}~\cite{luo2021separating}  first
separates text 
from the complex background  to make
TR easier.


\subsubsection{Spatial and Sequential Feature Extractors}
\label{ssec:rel:feaext}

\begin{figure}[ht]
	\centering
	\includegraphics[width=0.90\linewidth]{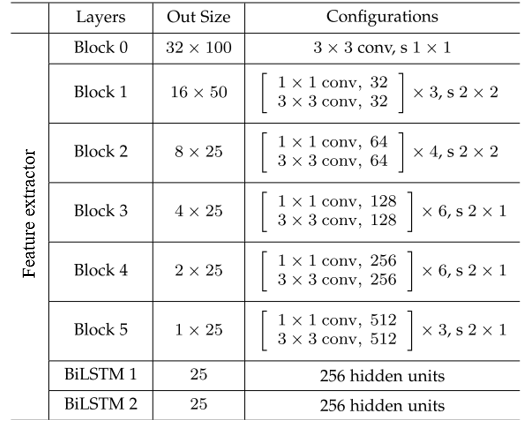}
	
	\vspace{-8px}
	\caption{Feature extractor in ASTER~\cite{shi2018aster}. 
		For a convolutional layer,
		``Out Size'' is the feature map size
		(height $\times$ width).
		For a sequential layer,
		``Out Size'' is 
		the
		sequence length. The symbol
		``s'' is the stride of the first convolutional layer in a block.
	}
	\label{fig:aster_extractor}
\end{figure}

Given an $H\times W$ input image,
the feature extractor~\cite{shi2018aster,shi2016end,yue2020robustscanner}
first uses a 
spatial model 
(which is a deep convolutional network)
to extract 
a $H^\prime\times W^\prime\times D$ feature map,
where
$H', W'$  are the downsampled height and width,
and
$D$ is the number of channels.
As an example,
consider 
the widely-used
ASTER~\cite{shi2018aster}.
Its pre-processing module
arranges the characters 
horizontally to a 
$32\times 100$ 
text image. 
Spatial features
are extracted 
by the ResNet~\cite{he2016deep}
(blocks 0-5 in Figure~\ref{fig:aster_extractor}).
Specifically,
2 sets of convolution filters (with stride $(2, 2)$) first downsample the image to
$8\times 25$, and then 3 sets of convolution filters (with stride $(2, 1)$)
further downsample it to a 
25-dimensional feature vector.
It also follows the common practice of 
doubling the number of convolution filters
when 
the 
feature map
resolution 
is changed
\cite{shi2016end,he2016deep,yang2019symmetry,baek2019wrong,wang2020decoupled}
(see also Figure~\ref{fig:search_space}).

The spatial model output is 
enhanced by 
extracting contextual information 
using a sequential model.
Specifically, its 
$H^\prime\times W^\prime\times D$ 
feature map 
is 
first reshaped to a 
$D \times T$
matrix,
where $T=H^\prime W^\prime$, and then 
processed
as
a sequence of $T$ 
feature vectors.
In ASTER,
BiLSTM~\cite{hochreiter1997long} layers
are built on top of the convolutional layers
(BiLSTM 1-2 in Figure~\ref{fig:aster_extractor}).

Design of the feature extractor in TR
is relatively less explored.
Often, 
they 
simply adopt existing architectures
\cite{shi2018aster,shi2016end,wang2020decoupled,yue2020robustscanner}.
Manual adjustments can be very time-consuming and expensive. 
Moreover, they 
do not consider the
latency constraints when 
the 
model is
deployed
on real devices 
(see Table~\ref{tab:txtreg}).

\subsubsection{Text Recognition~(TR) Head}

The TR head is used to recognize the text sequence.
In recent years,
many recognition heads have been proposed.
Connectionist temporal classification (CTC)~\cite{graves2006connectionist}
trains a classifier to match the prediction with the target text sequence without 
need for
prior alignment.
The segmentation-based TR head~\cite{liao2019scene,wan2020textscanner}
attempts to locate each character and 
then applies a character classifier.
Using the sequence-to-sequence model~\cite{SutskeverVL14},
the TR head in~\cite{shi2018aster,wang2020decoupled,yue2020robustscanner} 
uses the attention mechanism~\cite{LuongPM15} to
learn an implicit language model, which can be
parallelized
by
take the reading order as query
\cite{yu2020towards,wan2020textscanner}.
CTC and parallelized attention are more latency-friendly
than the sequence-to-sequence model, especially
when the output sequence is long.

\subsection{One-Shot Neural Architecture Search (NAS)} 
\label{sec:nas}


Traditionally, 
neural network architectures are taken as hyper-parameters,
and optimized by algorithms such as reinforcement learning~\cite{zoph2016neural} 
and evolutionary algorithms~\cite{xie2017genetic}. 
This is expensive as 
each candidate architecture 
needs to  be
fully
trained 
separately.
One-shot NAS
\cite{Cai19Proxylessnas,DARTS,pham2018enas}
significantly reduces the search time by sharing all network weights during training. 
Specifically, 
a supernet~\cite{pham2018enas,bender2018understanding,DARTS} 
subsumes all candidate architectures in the search space, and is trained only once. 
Each candidate architecture is a sub-network in the supernet, and its weights are simply inherited from the trained supernet
without training.

There are two approaches in
one-shot NAS.
The first one combines supernet training and search in a \textit{single} stage.
Representative methods include
DARTS~\cite{DARTS}, 
SNAS~\cite{xie2018snas}, 
ENAS~\cite{pham2018enas},
ProxylessNAS~\cite{Cai19Proxylessnas},
and
PNAS~\cite{Liu_2018_ECCV}.
A weight 
is used to reflect 
the importance of each candidate operator.
These weights 
are then learned 
together with the
parameters of
all candidate operators.
Finally, operators with higher weights are selected from the trained supernet to
construct the searched structure.
However, 
operators that do not perform well at the beginning 
of the training process
may not be fairly updated,
leading to 
the selection
of inferior architectures 
\cite{chu2019fairnas,li2020blockwisely}.
The second approach follows a \textit{two-stage} strategy,
which decouples supernet training and search.
In the pioneering work~\cite{bender2018understanding},
a large supernet
is trained
and 
sub-networks
are obtained 
by zeroing out some operators.
The best architecture is selected 
by measuring
the 
performance of
each sub-network.
Single-path one-shot (SPOS)~\cite{guo2020single} 
uniformly samples and updates a sub-network from the supernet in each iteration.
After the supernet has been sufficiently trained,
the best sub-network 
is selected
by an evolutionary algorithm.
Once-for-all (OFA)~\cite{CaiGWZH20} is similar to SPOS,
but
proposes a 
\textit{progressive shrinking}
scheme,
which 
trains
sub-networks 
in the supernet 
from large 
to small.
More recently,
DNA~\cite{li2020blockwisely} adopts knowledge distillation 
to improve fairness in supernet training,
and 
BossNAS further improve DNA by 
leveraging self-supervised learning~\cite{grill2020bootstrap}.



\label{sec:deploy}

On deploying deep networks to a specific device,
issues such as
model size
and 
inference time 
can become important. 
Above one-shot NAS 
methods have also been recently introduced to solve this problem.
For example,
MNASNet~\cite{tan2019mnasnet} and MobileNetV3~\cite{howard2019searching}
measure the actual execution latency of a sampled architecture, and use it as a 
reward to train a recurrent neural network controller.
ProxylessNAS~\cite{Cai19Proxylessnas} and FBNet~\cite{wu2019fbnet}
introduce a regularizer to the search objective
which measures the expected network latency (or number of parameters),
and stochastic gradient descent is used to search the architectures.
In SPOS~\cite{guo2020single},
an evolutionary algorithm is used to select architectures meeting the
deployment requirements.


\subsection{Transformer}
\label{sec:transformer}

For sequence modeling 
in natural language processing (NLP),
the LSTM 
has gradually been 
replaced
by the Transformer~\cite{VaswaniSPUJGKP17}, 
which is more
advantageous in allowing parallelism 
and 
extraction
of long-range context features. 
The transformer 
takes 
a 
length-$T$ 
sequence 
of
$D$-dimensional
features
$\mathbf{Z} \in \mathbb{R}^{D \times T}$
as input.
Using three multilayer perceptrons (MLPs),
$\mathbf{Z}$ is transformed 
to the query
$\mathbf{Q}$, key
$\mathbf{K}$,
and value $\mathbf{V}$, respectively
(all with size $D \times T$).
Self-attention (SA)  
generates attention scores  
$\mathbf{A}$
from
$\mathbf{Q}$ and 
$\mathbf{K}$:
\begin{align}
\mathbf{A} = \mathbf{Q}^{\top} \mathbf{K} / \sqrt{D},
\label{eq:tranatt}
\end{align}
where $\sqrt{D}$ is a scaling factor, 
and then use
$\mathbf{A}$ to form
weighted sums of columns in $\mathbf{V}$:
\begin{align*}
\text{SA}(\mathbf{Q}, \mathbf{K}, \mathbf{V}) 
= \mathbf{V} \cdot\text{softmax}(\mathbf{A}),
\end{align*}
where 
$\left[ \text{softmax}(\mathbf{A}) \right]_{i,j}=e^{\mathbf{A}_{i,j}} / \sum_{t=1}^{T} e^{\mathbf{A}_{i, t}}$.
Multiple attention heads
$\{(\mathbf{Q}_i, \mathbf{K}_i, \mathbf{V}_i)\}$ 
can also be used, leading to 
multi-head self-attention (\textit{MHSA})~\cite{VaswaniSPUJGKP17}:
\begin{align*}
\textit{MHSA}\left(\mathbf{Q}, \mathbf{K}, \mathbf{V}\right) 
= \text{Concat}
\left( 
\text{head}_1, 
\dots,
\text{head}_I
\right)  \mathbf{W}^O,
\end{align*}
where $ \text{head}_i = \text{SA}( \mathbf{Q}_i, \mathbf{K}_i,  \mathbf{V}_i) $,
$\mathbf{W}^O$
is a learnable parameter,
and
$\text{Concat}(\dots)$ concatenates multiple column vectors to 
a single vector.
Finally, the \textit{MHSA} output is followed by a two-layer
feedforward network (\textit{FFN}):
\begin{equation} \label{eq:mlp}
\text{MLP}\left(\mathbf{x}\right)=\text{ReLU}\left(\mathbf{x} \mathbf{W}_1\right) \mathbf{W}_2,
\end{equation} 
where $\mathbf{x} = \textit{MHSA}(\mathbf{Q}, \mathbf{K}, \mathbf{V})$,
$\mathbf{W}_1, \mathbf{W}_2$ 
are learnable parameters, 
and ReLU is the rectified linear activation function.

With the initial success of the transformer, a lot of 
efforts have been made to 
improve
its
two core components:
\textit{MHSA} and \textit{FFN}.
The RealFormer~\cite{he2020realformer}
adds a skip-connection in
\textit{MHSA} 
to help propagation of the raw attention scores
and 
stabilize
training.
The addition of relative distance awareness information 
can also improve
self-attention 
\cite{DaiYYCLS19}.
Besides, there are efforts 
to improve
the computational efficiency of \textit{MHSA} (as in
Reformer~\cite{kitaev2020reformer} and
Performer~\cite{choromanski2020rethinking}). 
As for \textit{FFN} component, 
the Evolved Transformer~\cite{So2019TheET}
uses NAS to find a better \textit{FFN} structure.

\begin{figure*}[t]
	\centering
	{\includegraphics[width = 0.90\textwidth]{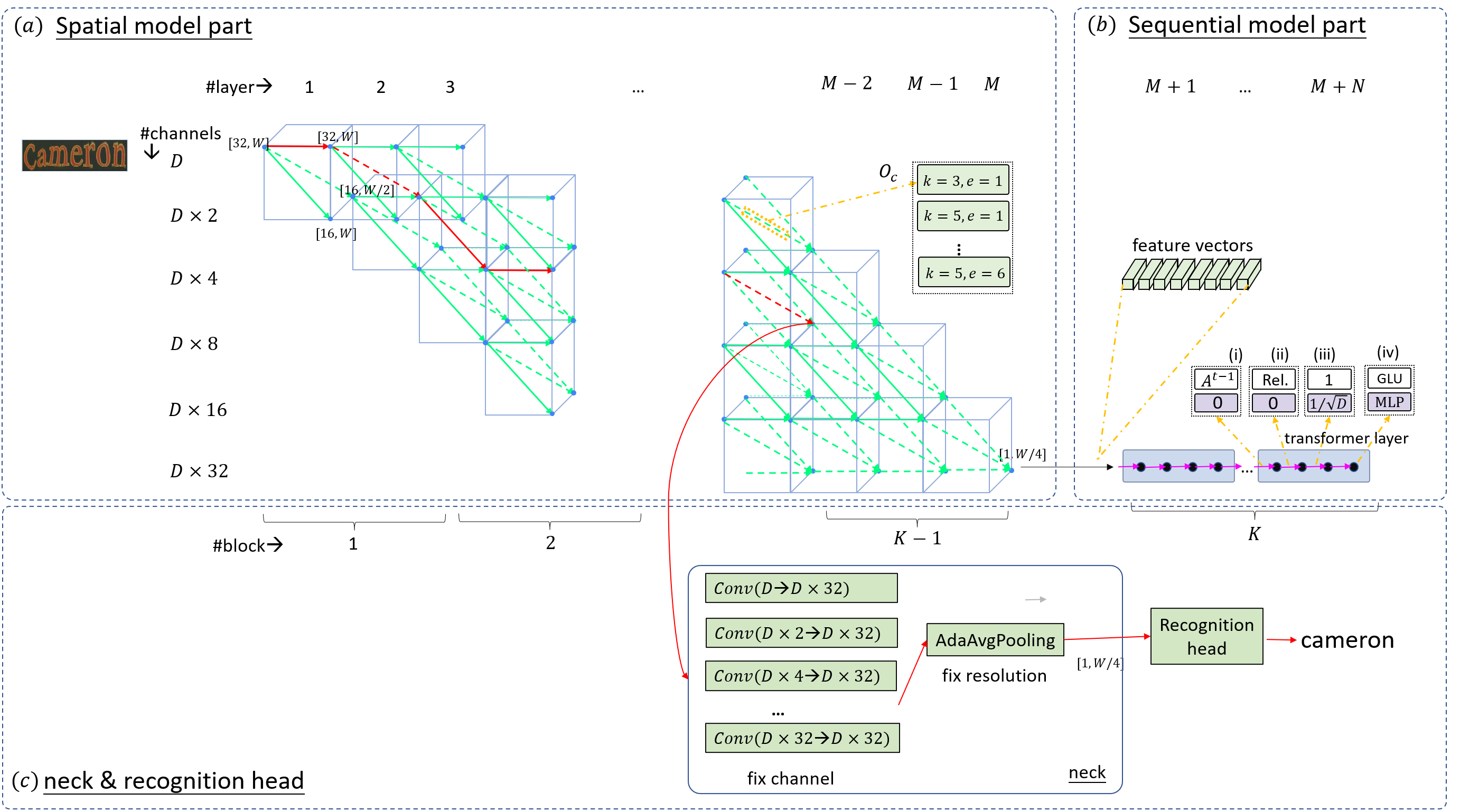}}
	\vspace{-10px}
	\caption{
		Graph illustrate of the proposed method TREFE.
		The search space of TREFE contains both spatial model (see Section~\protect\ref{sec:path_search_space}) and sequential model (see Section~\protect\ref{sec:rnn_search_space}) part,
		and the neck is used for supernet training (see Section~\protect\ref{ssec:train_supernet}).}
	\label{fig:search_space}
	\vspace{-10px}
\end{figure*}



\section{Proposed Methodology}

As discussed in 
Section~\ref{ssec:rel:feaext},
the feature extractor 
has two components: 
(i) a spatial model for visual feature extraction,
and (ii) a sequential model for sequence feature extraction.
In TR,
the feature extractor 
design 
is not well studied.
Existing TR algorithms often simply
reuse spatial and sequential model architectures that are originally designed for other tasks
\cite{shi2018aster,shi2016end,wang2020decoupled,yue2020robustscanner}.
However, it has been recently observed that 
different tasks
(such as semantic segmentation~\cite{liu2019auto} and object detection~\cite{chen2019detnas})
may require  different
feature extraction
architectures.
Thus, existing
spatial and sequential feature extractors may not be
suitable for TR.
On the other hand, manual adjustments can be very time-consuming and expensive. 
Moreover, 
when the model is deployed on real devices,
latency constraints 
cannot be easily considered.

Inspired by NAS,
we propose the
TREFE algorithm,
which automatically searches for a 
high-performance \underline{T}ext \underline{RE}cognition \underline{F}eature \underline{E}xtractor.


\subsection{Formulating TR as a NAS Problem}
\label{search_space}

In this section, 
we first formulate 
TR
as a 
NAS 
problem.
The search space is
inspired by existing manually-designed TR models 
and recent advances in the transformer architecture.


\subsubsection{Search Space for the Spatial Model}
\label{sec:path_search_space}

Recall that the 
spatial model  is a CNN
(Section~\ref{ssec:rel:feaext}).
Each convolutional layer $\M{C}$ can be represented as 
$\M{C}(\mathsf{X}; ct, s^h, s^w)$,
where $\mathsf{X}$ is the input image tensor, 
$ct$ is the type of convolution 
(e.g., a 3$\times$3 convolution or 5$\times$5 depth-wise separable convolution),
and
$(s^h, s^w)$ are the 
strides in the
height and width dimensions.
The downsampling path, which downsamples  the image to the feature map 
along 
with the convolution operations,
can significantly affect
the CNN's 
performance
\cite{liu2019auto,shi2016end,chen2019detnas}.
Instead of using manual designs,
we explore 
the 
space
of spatial model architectures
and automatically search for the 
($ct, s^h,s^w$) values in each layer.

The whole spatial model structure 
can be identified by 
$C \equiv \{ ct_i \}_{i = 1}^M$ 
and $S \equiv \{(s^h_i, s^w_i)\}_{i = 1}^M$,
where 
$M$ is the number of convolution layers, 
$ct_i \in O_c$, the set of candidate convolution operations, and
$(s^h_i, s^w_i) \in O_s$, the set of candidate stride values.
Following~\cite{wang2020decoupled,cheng2017focusing,shi2018aster}, we assume that the input to the spatial model is of size $32\times W$.
The detailed choices of $O_s$ and $O_c$ are as follows.
\begin{itemize}[leftmargin=*]
\item 
Following~\cite{shi2018aster,yang2019symmetry,wang2020decoupled,bai2018edit,shi2016end},
we set
$O_s =\{(2, 2), (2, 1), (1, 1)\}$. 
We do not include $(1, 2)$,
as 
the horizontal direction 
should not
be downsampled more 
than the vertical
direction~\cite{bai2018edit,shi2016end,baek2019wrong,yang2019symmetry,zhan2019esir},
as this can 
make neighboring characters more difficult to 
separate.
Moreover, we
double
the number of filters 
when the resolution in that layer is
reduced (i.e., when $(s^h_i, s^w_i)= (2, 1)$ or $(2, 2)$).

\item As in NAS algorithms~\cite{wu2019fbnet,Cai19Proxylessnas,hong2020memory,howard2019searching},
$O_c$ contains
inverted bottleneck convolution 
(MBConv) layers~\cite{sandler2018mobilenetv2}
with kernel size $k$ $\in$ $\{3$, $5\}$  and
expansion factor $e$ $\in$ $\{1$, $6\}$.

\item As in~\cite{shi2016end,shi2018aster,zhang2020autostr}, 
we use
an output feature map of size $1\times W/4$.
Since the input size of the spatial model is $32\times W$,
we have
for each downsampling path, 
\begin{align}
S^h\equiv
32
= 
\prod\nolimits_{i=1}^{M}s^h_i,
\quad\text{and}\quad
S^w \equiv
4=
\prod\nolimits_{i=1}^{M}s^w_i.
\label{eq:size}
\end{align}
\end{itemize}

Figure~\ref{fig:search_space}(a)
shows the search space of
the spatial model structure
with $M$ layers.
Each blue node corresponds to a
$h\times w\times c$ 
feature map
$\bm{\beta}_{(h,w,c)}^l$ at layer $l$.\footnote{In general, the $(h,w,c)$  values can vary with $l$.}
Each green edge
corresponds to a candidate
convolution layer $\mathcal{C}$ transforming
$\bm{\beta}_{(h,w,c)}^l$, while
each gray edge 
corresponds to a candidate
stride in $O_s$.
A connected path
of blue nodes from  the initial size ($[32,W]$) to the size 
of the last
feature map 
($[1,W/4]$)
represents 
a candidate spatial model.


\subsubsection{Search Space for the Sequential Model}
\label{sec:rnn_search_space}

Recall 
that 
in TR systems,
the sequential model component 
is usually a
recurrent network
(Section~\ref{ssec:rel:feaext}),
such as the BiLSTM~\cite{hochreiter1997long}.
Here, we 
instead
use the transformer~\cite{VaswaniSPUJGKP17}
which has higher parallelism.
However, a straightforward application of the
vanilla Transformer may not be desirable, as
it can have performance inferior to the BiLSTM
on tasks such as
named entity recognition~\cite{yan2019tener} and 
natural language inference~\cite{GuoQLSXZ19}.  
In the following, we describe the proposed chanages to the
Transformer structure.

\begin{figure}[ht]
	\centering
	\includegraphics[width=0.8\columnwidth]{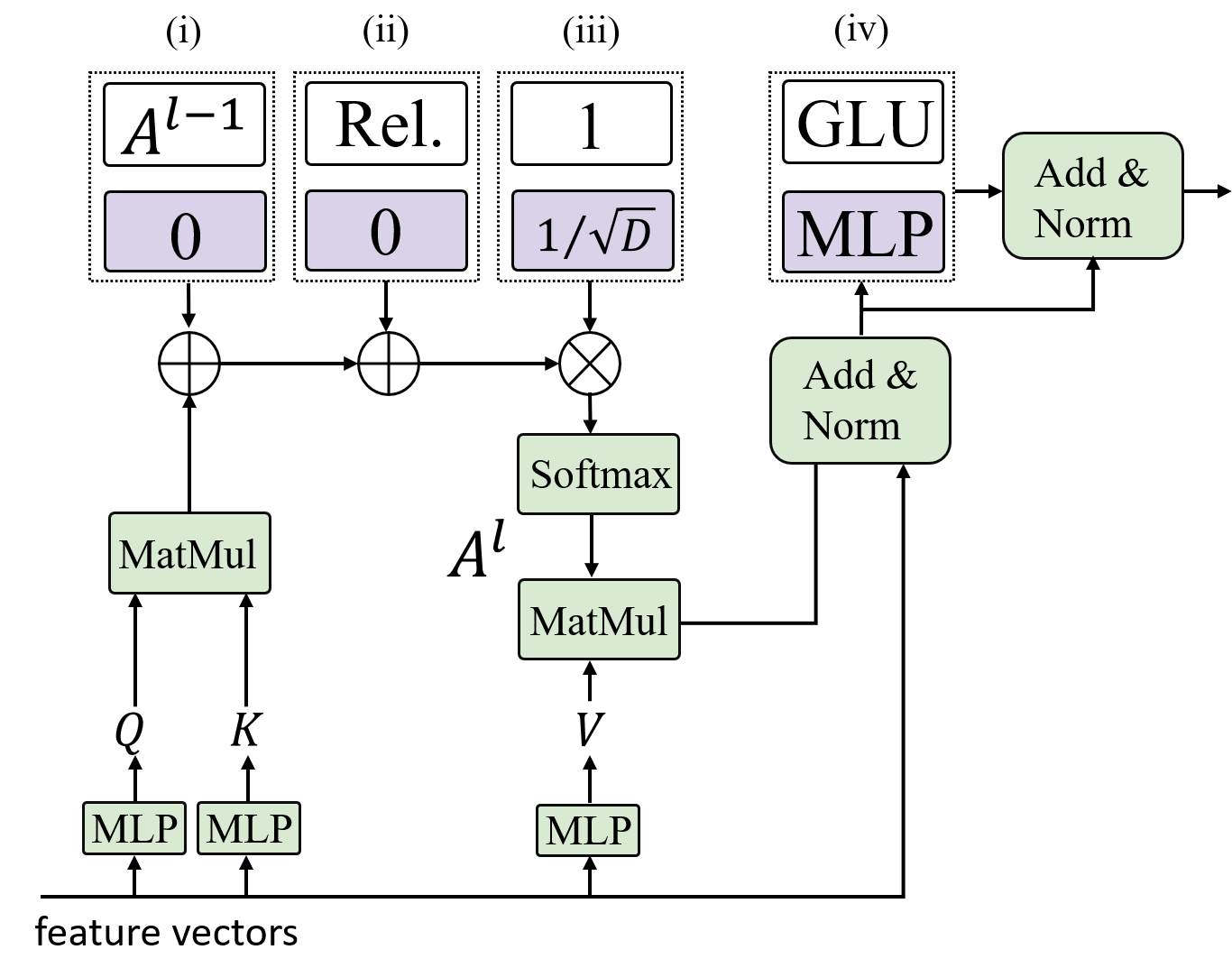}
	
	\vspace{-10px}
	\caption{Search space of a transformer layer.
	``Rel.'' is short for relative distance embedding.
	The purple boxes represent the original structures, and 
	boxes with white background indicates the alternative choices.}
	\label{fig:transformer_search_space}
	\vspace{-10px}
\end{figure}

Let each transform layer be $\M{R}(\mathsf{V}, rt)$,  
where
$\mathsf{V}$ is the input tensor and $rt$ is the type of transform layer 
(e.g., a transformer layer without attention scaling).
The structure of the sequential model is defined as
$R \equiv \{ rt_i \}_{i = 1}^N$ where $rt_i \in O_r$, 
the set of candidate layers, and $N$ is the number of transformer layers.
Consider the
$\ell$th
transform layer.
Inspired by
recent advances on the transformer (Section~\ref{sec:transformer}), 
one can vary its design in the following four aspects. 
The first three are related to the 
\textit{MHSA}, while the last one is on the
\textit{FFN}
(Figure~\ref{fig:transformer_search_space}).

\begin{itemize}
\item[(i)] 
As in the RealFormer~\cite{he2020realformer},
one can add a residual
path from the previous layer to the current layer to facilitate
propagation of attention scores. In other words,
instead of using $\mathbf{A}^\ell= \mathbf{Q}^{\top} \mathbf{K}$, we can have
\begin{align*}
\mathbf{A}^\ell = \mathbf{Q}^{\top} \mathbf{K} + \mathbf{A}^{\ell - 1},
\end{align*}
for transform layer $\ell>1$ (excluding the input layer).

\item[(ii)] 
As in
\cite{DaiYYCLS19,yan2019tener}, 
one can add a relative distance embedding 
$\textsf{Rel}$ 
to
improve the distance and direction awareness of
the attention score $\mathbf{A}$ in \eqref{eq:tranatt} as:
\[ 
\mathbf{A}
= 
\mathbf{A} + \textsf{Rel},
\]
where
$\textsf{Rel}_{tj} = \mathbf{q}_t \mathbf{r}^{\top} + \mathbf{u}
\mathbf{k}_{j}^{\top} + \mathbf{v} \mathbf{r}^{\top}$,
$\mathbf{u}, \mathbf{v}$ are learnable parameters,
$\mathbf{q}_t$ is the 
$t$th query ($t$th column in $\mathbf{Q}$),
$\mathbf{k}_j$ is the $j$th key ($j$th column in $\mathbf{K}$),
$\mathbf{r}=[
\mathbf{r}_{i}]
$ is the
relative position between 
$\mathbf{q}_t$ and  $\mathbf{k}_j$, and is defined as
\begin{align*}
\mathbf{r}_{i}
= 
\begin{cases}
\sin\big( (t-j) / ( 10000^{\frac{i}{D}} ) \big) 
\! & \! i \text{ is even}
\\
\cos\big( (t-j) / ( 10000^{\frac{i - 1}{D}} ) \big) 
\! & \! i \text{ is odd}
\end{cases}.
\end{align*}

\item[(iii)] 
In computing the attention
in (\ref{eq:tranatt}),
instead of including the scaling factor $1/\sqrt{D}$,
one can drop
this
as in~\cite{yan2019tener}, leading to simply
$\mathbf{A} = \mathbf{Q}^{\top} \mathbf{K}$.

\item[(iv)] 
Inspired by~\cite{yan2021primitive,dauphin2017language}, 
one can replace the 
\textit{FFN}'s
MLP in \eqref{eq:mlp} 
with the gated linear unit (GLU)~\cite{dauphin2017language}:
\begin{equation} \label{eq:glu}
\text{GLU}(\mathbf{x})=(\mathbf{x} \mathbf{W}_1) \otimes \sigma(\mathbf{x}  \mathbf{W}_2),
\end{equation}
where $\mathbf{x}$ is the 
\textit{MHSA}
output,
$\otimes$ is the element-wise product, and $\sigma$ is the sigmoid function.
This allows the \textit{FFN} to select relevant features  for prediction.
\end{itemize}
For a sequential model
with $N$ layers, we  attach $N$ copies of 
Figure~\ref{fig:transformer_search_space} to the sequential model  output in
Figure~\ref{fig:search_space}(a).
In Figure~\ref{fig:search_space}(b),
each blue rectangle denotes a transformer layer, and
each black node 
denotes the design choices
in the transformer layer 
((i), (ii), (iii), (iv) in
Figure~\protect\ref{fig:transformer_search_space}).
The 
(magenta)
path 
(with color magenta)
in Figure~\ref{fig:search_space}(b) along the black nodes 
constructs a candidate
sequential model.


\subsubsection{Resource Constraints} \label{ssec:resource_constrain}

As in Section~\ref{sec:deploy},
there can be resource constraints 
on deployment.
We introduce resource constraints  of the form:
\begin{equation}
\text{latency}
\big(
\M{N}(\mathbf{w}, S, C, R); \mathcal{E}
\big) 
\le r_{\max},
\label{eq:latent}
\end{equation}
where $\M{N}(\mathbf{w}, S, C, R)$ is the TR model with network
weights $\mathbf{w}$ and feature extractor architecture determined by $(S, C,
R)$, 
$\mathcal{E}$ is the environment,
and $r_{\max}$ is the budget 
on the resource.
For simplicity, only one resource constraint is considered. Extension to multiple resource constraints is straightforward.


\subsubsection{Search Problem}
\label{ssec:search_problem}

Consider a feature extractor 
with $M$ layers 
in the spatial model and $N$ in the sequential model.
Let $\M{L}_{\text{tra}}$ be the 
training loss of the network,
and $\M{A}_{\text{val}}$ be the quality of the network
	on the validation set.
The following formulates the search for an appropriate architecture $\M{N}$.

\begin{definition} \label{def:pro}
The search
problem can be formulated as:
\begin{align}
& \arg\max\nolimits_{S, C, R}
\mathcal{A}_{\text{val}}(\M{N}(\mathbf{w}^*, S, C, R))
\notag
\\
& \text{s.t.}
\begin{cases}
	\mathbf{w}^* 
	= \arg\min_{\mathbf{w}}
	\mathcal{L}_{\text{tra}}(\M{N}(\mathbf{w}, S, C, R)),
	\\
	\text{latency}\big(\M{N}(\mathbf{w}, S, C, R); \mathcal{E} \big) \le r_{\max},
	\\ 
	C \in \mathbb{C},
	\quad
	R \in \mathbb{R},
	\quad
	S \in \mathbb{P},
\end{cases}
\!\!\!\!
\label{eq:opt}
\end{align}  
where
\[ \mathbb{C} 
\equiv \underbrace{O_c \times \cdots \times O_c}_{M},
\; \mathbb{R} \underbrace{\equiv O_r \times \cdots \times O_r}_{N},
\]
and
$\mathbb{P}
\equiv
\{
\{ (s^h_i, s^w_i) 
\in O_s 
\}_{i = 1}^M
\;|\;
\prod\nolimits_{i=1}^{M}s^h_i \! = \! S^h, 
\prod\nolimits_{i=1}^{M}s^w_i \! = \! S^w
\}$
encodes the constraints in  (\ref{eq:size}).
\end{definition}

The combination of 
$(S, C, R)$ generates a large number of candidate architectures.
For example,
in the experiments (Section~\ref{sec:exp}),
we use
a spatial model 
with 20 MBConv layers. 
Every blue node in Figure~\ref{fig:search_space} 
has 3 
stride $(s_h, s_w)$ 
candidates 
and 
4 
operator ($O_c$)
candidates.
There are a total of 155,040
candidate downsampling paths,\footnote{
A backtracking algorithm 
computes the number of candidate downsampling paths. 
Please refer to Appendix~\ref{app:size} for details.}
$155,040\times4^{20} \approx  1.7\times10^{17}$
candidate spatial model structures; whereas
the sequential model has
4 transformer layers, and
$(2^4)^4=65,536$
candidate structures. The whole search space thus has
$1.7\times10^{17} \times 65,536 \approx 1.1\times10^{22}$ candidates,
which is prohibitively large.
Besides, note that problem \eqref{eq:opt} 
is a bi-level optimization problem. As in most NAS
problems~\cite{elsken2019neural},
it is typically expensive to solve.
In particular, training each candidate architecture to get $\mathbf{w}^*$ is expensive.
Hence, directly optimizing problem \eqref{eq:opt} is not practical.


\subsection{Search Algorithm}
\label{sec:search_alg}

Inspired by recent advances in NAS~\cite{elsken2019neural},
we propose to solve \eqref{eq:opt} 
using one-shot NAS~\cite{guo2020single,DARTS,Cai19Proxylessnas},
which greatly simplifies the search by training only one supernet.
However, the one-stage 
approach 
requires
training the whole supernet,
which demands tremendous GPU memory as the proposed search space is huge 
\cite{Cai19Proxylessnas}.
Hence, we will use the two-stage approach.
However,  
the 
two-stage 
methods 
cannot be directly used.
As the search space is huge,
only 
a small fraction of the candidate architectures  can be
trained,
and the untrained architectures will perform badly
\cite{li2020blockwisely}.

In Section~\ref{sec:design_supernet},
we first discuss
design of the supernet.
Inspired by layer-wise pre-training~\cite{bengio2007greedy,hinton2006fast},
we propose 
in Section~\ref{ssec:train_supernet}
a progressive strategy 
that trains the supernet
in a block-wise manner.
In Section~\ref{ssec:search},
we propose to use natural gradient descent~\cite{amari1998natural} to better search for a sub-architecture
during the second stage.


\subsubsection{Designing the Supernet
in One-Shot NAS}
\label{sec:design_supernet}
There are two basic requirements in 
the supernet design
\cite{bender2018understanding,pham2018enas,DARTS}:
(i) all candidates in the search space should be included; and
(ii) each candidate can be expressed as a path in the supernet.

The proposed supernet has two parts: the spatial model and 
sequential model.
It closely matches the search space in
Figure~\ref{fig:search_space}.
The spatial component
(Figure~\ref{fig:search_space}(a))
is a 3D-mesh,
in which each edge 
determines an operation that transforms the feature representation.
A connected path from $[32, W]$ to $[1, W/4]$ represents a downsampling path.
The choice of operations and downsampling path together determine the
CNN.
Figure~\ref{fig:search_space}(b)
shows
the sequential model component of the supernet.


\subsubsection{Training the Supernet}
\label{ssec:train_supernet}

The main challenge 
is how to fully and fairly train all candidate architectures in 
the supernet.
A typical solution is to
sample architectures uniformly and then train 
\cite{li2019random,guo2020single}.
However, uniform sampling in a huge search space 
is not effective.
To alleviate this problem,
we propose to divide the supernet 
(denoted $\Phi$)
into $K$ smaller blocks
($\Phi_1, \Phi_2, \dots, \Phi_{K}$)
and 
optimize them one by one.
Since
the spatial model is much larger than the sequential model,
we take the whole sequential model 
as one block
($\Phi_{K}$),
and divide the spatial model into $K-1$ equal-sized blocks 
($\Phi_1, \Phi_2, \dots, \Phi_{K-1}$)
(Figure~\ref{fig:search_space}(a)).

Algorithm~\ref{alg:prog} shows the training process.
The weights for blocks
$\Phi_{1},\dots,\Phi_{K}$
are progressively stacked, 
and updated together with
the weights of the neck and recognition head by SGD.
When training block $\Phi_k$,
we fix the trained weights for blocks
$\Phi_1,\dots, \Phi_{k - 1}$, and
skip the remaining blocks $\Phi_{k + 1},\dots, \Phi_K$ 
(step~\ref{alg:prog:fix}).
In each iteration,
a path 
$\bm{\alpha}_k$ 
is uniformly sampled from $\Phi_k$
(step~\ref{alg:prog:alphak}).
Let $\mathbb{S}_k$ be the set of paths in $\Phi_1\cup \dots \cup \Phi_{k - 1}$
whose output feature maps 
match in
size  with
the input feature map of $\bm{\alpha}_k$.
A new path in $\Phi_1\cup \dots \cup \Phi_{k}$ is formed 
by uniformly selecting a path $\bm{\alpha}_k^{\leftarrow} \in \mathbb{S}_k$ and
connecting it with $\bm{\alpha}_k$
(step~\ref{alg:prog:alphak2}).
As blocks $\Phi_{k + 1},\dots, \Phi_K$ are
skipped, 
the output of $\bm{\alpha}_k$ is connected to the recognition head
via an extra auxiliary neck\footnote{In the experiments, the neck is a small network with six parallel convolution layers and a adaptive pooling layer\cite{he2016deep}.}
(step~\ref{alg:prog:head})
so that
$\bm{\alpha}_k$'s 
output 
channel number 
and resolution 
match 
with
those of the head's input.

\begin{algorithm}[ht]
\caption{Training the supernet.}
\small
\begin{algorithmic}[1]
	\State Split the supernet into $K$ blocks;
	
	\State Insert an auxiliary neck between feature extractor and recognition head; 
	
	\For{block $k = 1, \dots, K$}
	\State fix supernet weights for blocks $\Phi_1,\dots,\Phi_{k-1}$;
	\label{alg:prog:fix}
	
	\For{iteration $t = 1, \dots, T$}
	
	\State sample a path $\bm{\alpha}_k$ from $\Phi_k$;
	\label{alg:prog:alphak}
	
	\State sample a path $\bm{\alpha}_{k}^{\leftarrow}$ in $\mathbb{S}_k$
	and connect it with $\bm{\alpha}_k$;
	\label{alg:prog:alphak2}
	
	\State sample a mini-batch $B_t$ from training data;
	
	\State update weights of 
	$\bm{\alpha_k}$, the neck and the recognition head 
	by SGD on $B_t$; 
	\label{alg:prog:head}
	
	\EndFor
	\EndFor
	
	\State \textbf{return} trained weights $\mathbf{W}^*$ of $\Phi$.
\end{algorithmic}
\label{alg:prog}
\end{algorithm}

\begin{table*}[ht]
	\centering
	\caption{Comparison between AutoSTR~\cite{zhang2020autostr} 
	and TREFE. Here, ``SeqAtt" denotes sequential attention, and
	``ParAtt" denotes parallel attention.}
	\label{tab:comp}
	\vspace{-10px}
	\begin{tabular}{c | c | c | c | c | c | c | c}
		\toprule
		&       \multicolumn{2}{c}{spatial model}       & sequential           & \multicolumn{2}{c}{recognition head} & search    & deployment \\
		& downsampling path     & operators             & model                & scene text & handwritten text        & algorithm & awareness  \\ \midrule
		AutoSTR                        & limited choices & \multirow{2}{*}{same} & fixed BiLSTM         & SeqAtt     & ---                     &  decoupled search         &     not support       \\		\cmidrule{1-2}\cmidrule{4-8}
		TREFE                           & all possible paths    &                       & searched transformer & ParAtt     & CTC                     &      joint     search &    support        \\ \bottomrule
	\end{tabular}
	\vspace{-10px}
\end{table*}


\subsubsection{Search for a Sub-Network}
\label{ssec:search}

Recall that a 
path 
$\bm{\alpha}$ 
in the supernet 
corresponds to an architecture 
$\{ S, C, R \}$
of the feature extractor.
Let 
the trained weight 
of the supernet 
returned from Algorithm~\ref{alg:prog} be $\mathbf{W}^*$.
Since the constraints
$C \in \mathbb{C}$,
$R \in \mathbb{R}$,
and
$S \in \mathbb{P}$
have been implicitly encoded by the supernet structure,
and the supernet weights are already trained,
problem~\eqref{eq:opt} then simplifies to
\begin{align}
\bm{\alpha}^* & = \arg\max\nolimits_{\bm{\alpha}}
\mathcal{A}_{\text{val}}
\big(
\M{N}(\mathbf{W}^*(\bm{\alpha}), \bm{\alpha})
\big)
\notag
\\
\text{s.t.} &
\quad
\text{latency}\big(\M{N}(\mathbf{W}^*(\bm{\alpha}), \bm{\alpha}); \mathcal{E} \big) \le r_{\max},
\label{eq:search_on_supernet}
\end{align} 
where 
$\mathbf{W}^*(\bm{\alpha})$  is the weight for path
$\bm{\alpha}$,
which can be easily extracted from
$\mathbf{W}^*$ without re-training.
SPOS~\cite{guo2020single} 
uses an evolutionary algorithm (EA)
to solve \eqref{eq:search_on_supernet}.
However, 
EA can suffer from the pre-maturity problem~\cite{LeungGX97}, in that
the population is dominated by good architectures generated in the early stages.
Diversity is rapidly reduced, and
the EA converges to a locally optimal solution.

To avoid the above problem, 
we consider using
stochastic relaxation on $\bm{\alpha}$ as in~\cite{xie2018snas},
and transform problem \eqref{eq:search_on_supernet} to:
\begin{align}
	\max\nolimits_{\bm{\theta}}
	\; & \; 
	\mathbb{E}_{\bm{\alpha} \sim P_{\bm{\theta}}(\Phi) } 
	\left[\mathcal{A}_{\text{val}}
	\big(
	\M{N}(\mathbf{W}^*(\bm{\alpha}), \bm{\alpha})
	\big)\right]
	\nonumber\\
	\text{s.t.}
	&
	\quad
	\text{latency}\big(\M{N}(\mathbf{W}^*(\bm{\alpha}), \bm{\alpha}); \mathcal{E} \big) \le r_{\max},
	\label{eq:search_on_supernet_ng}
\end{align}
where
$\mathbb{E}$ denotes the expectation,
$P_{\bm{\theta}}$ is an exponential distribution (with parameter
$\theta$)
on the search space $\Phi$
(details are in Appendix~\ref{app:dist}).
Sampling from $P_{\bm{\theta}}$ helps
to explore more diverse architectures.

Algorithm~\ref{alg:search_supernet} shows the search procedure.
To optimize $\bm{\theta}$, 
we first sample 
a mini-batch $B$ of architectures 
using the exponential distribution $P_{\bm{\theta}}$
(step~5).
For each sampled architecture,
its latency  and validation set performance 
are measured (steps~7-8).
Note that 
this 
takes negligible time compared to
supernet
training.
Architectures that do not meet the latency requirements  are dropped.
The sampled architectures and corresponding performance scores are used to update $P_{\bm{\theta}}$
by natural gradient descent~\cite{amari1998natural}
(steps~18).
Specifically,
at the $t$th iteration, 
$\bm{\theta}$ is updated as:
\begin{align}
&\bm{\theta}_{t + 1} = \bm{\theta}_t + \rho \mathbf{F}^{-1} (\bm{\theta}_t) 
\mathbf{g},
\label{eq:ngd} 
\end{align}
where $\rho$ is the step-size,
\begin{align}
\mathbf{F} (\bm{\theta}) = \mathbb{E}_{P_{\bm{\theta}}} 
\left[ 
\nabla_{\bm{\theta}}\ln P_{\bm{\theta}}(\bm{\alpha}) \left[\nabla_{\bm{\theta}}\ln P_{\bm{\theta}}(\bm{\alpha})\right]^\top
\right] ,
\label{eq:fisher}
\end{align}
is the Fisher information matrix~\cite{amari1998natural},
and 
$\mathbf{g}$ is
the gradient 
\begin{align}
\mathbf{g} =
\mathbb{E}_{P_{\bm{\theta}}} 
\Big[
\mathcal{A}_{\text{val}}(\M{N}(\mathbf{W}^*(\bm{\alpha}), \bm{\alpha})) \nabla_{\bm{\theta}} \ln \left( P_{\bm{\theta}} \left(\bm{\alpha}\right) \right) 
\Big].
\label{eq:grad}
\end{align}
Note that 
$\mathbf{F} (\bm{\theta})$ and
$\mathbf{g}$ in
\eqref{eq:fisher},
\eqref{eq:grad} cannot be exactly evaluated as they require
expensive integrations over the whole distribution.
Thus, 
they are approximated
by averages over the sampled architectures
(steps~14-15).
Finally,
step~20 returns
the architecture with
the best validation performance
(that also satisfies the latency requirement).
Finally,
Algorithm~\ref{alg:whole_alg} shows the whole training procedure for TREFE.


\begin{algorithm}[ht]
	\caption{Search for a sub-network.}
	\small
	\begin{algorithmic}[1]
		\State $\bm{\alpha}^* \gets \emptyset, \text{perf}^* \gets -\inf $;
		\For{iteration $t$ = 1 \textbf{to} $T$}
		\State $j = 0, \mathbf{g} = 0$, $\mathbf{F} (\bm{\theta}) \leftarrow \mathbf{0}$;
		\While{$j < B$}
		\State sample $\bm{\alpha} \sim P_{\bm{\theta}} \left(\Phi\right)$;
		\State obtain network weight $\mathbf{w}^* \leftarrow \mathbf{W}^*(\bm{\alpha})$;
		\State $r \leftarrow \text{latency}\big(\M{N}(\mathbf{w}^*, S, C, R); \mathcal{E} \big)$;
		\State $\text{perf} \leftarrow \mathcal{A}_{\text{val}}(\M{N}(\mathbf{w}^*, S, C, R))$;
		\If{$r \le r_{\max}$} 
		\If {$\text{perf} > \text{perf}^*$}
		\State $(\bm{\alpha}^*, \text{perf}^*) \leftarrow (\bm{\alpha}, \text{perf})$;
		\EndIf 
		\EndIf
		\State $\mathbf{F} (\bm{\theta})
		\!\leftarrow\! 
		\big(
		j  \mathbf{F} (\bm{\theta}) 
		\! + \!  
		\nabla_{\bm{\theta}}\ln P_{\bm{\theta}}(\bm{\alpha}) \left[\nabla_{\bm{\theta}}\ln P_{\bm{\theta}}(\bm{\alpha})\right]^\top
		\!
		\big)
		/
		(j \! + \! 1)$;
		\State $\mathbf{g}
		\!
		\leftarrow
		\!
		\big(
		j \mathbf{g} 
		\! + \! 
		\mathcal{A}_{\text{val}}(\M{N}(\mathbf{W}^*(\bm{\alpha}), \bm{\alpha})) \nabla_{\bm{\theta}} \ln \left( P_{\bm{\theta}} \left(\bm{\alpha}\right) \right)
		\big) 
		/ (j \! + \! 1)$;
		\State $j \leftarrow j + 1$; \label{alg:k}
		\EndWhile
		\State update $\bm{\theta}$ via \eqref{eq:ngd} using \eqref{eq:fisher} and \eqref{eq:grad};
		\EndFor
		\State \textbf{return} searched architecture $\bm{\alpha}^*$.
	\end{algorithmic}
	\label{alg:search_supernet}
\end{algorithm}

\begin{algorithm}[ht]
	\caption{\underline{T}ext \underline{RE}cognition \underline{F}eature \underline{E}xtractor (TREFE).}
	\small
	\begin{algorithmic}[1]
		\State Build a supernet $\Phi$ (see Section~\ref{sec:design_supernet});
		\State Train $\Phi$ progressively on training data via Algorithm~\ref{alg:prog};
		\State Search $\bm{\alpha}^*$ from $\Phi$ on validation data via Algorithm~\ref{alg:search_supernet};
		\State Re-train the $\bm{\alpha}^*$ from scratch;
	\end{algorithmic}
	\label{alg:whole_alg}
\end{algorithm}

\subsection{Comparison with AutoSTR}
\label{sec:diff:autostr}

There are several differences between 
AutoSTR~\cite{zhang2020autostr} and
the proposed TREFE 
(Table~\ref{tab:comp}):
\begin{enumerate}
\item 
Search algorithm:
AutoSTR only 
searches the spatial model, and
the
downsampling paths 
and
operators  
are searched
separately
(using grid search and
ProxylessNAS, respectively).
On the other hand,
TREFE jointly searches both the spatial and sequential models (including the
downsampling paths, operators and transformer architecture).

\item
To make AutoSTR efficient,
only 10 types
of downsampling paths are allowed during the search
(step~1 in Section~3 of~\cite{zhang2020autostr}).
On the other hand,
by using the progressive training strategy,
TREFE can efficiently explore all possible downsampling paths by sharing weights in a supernet.
Thus, combined with the joint search process,
TREFE
can find better spatial and sequential models than AutoSTR.

\item 
As in ASTER~\cite{shi2018aster}, 
AutoSTR uses
a sequential attention-based sequence-to-sequence decoder~\cite{BahdanauCB14} as the recognition head.
As characters are output
one-by-one, 
this is not latency-friendly,
especially when the output text sequence is long 
(as in handwritten text).
In contrast, 
TREFE outputs the text sequence in parallel by
using parallel attention~\cite{yue2020robustscanner,yu2020towards} 
(resp. CTC head~\cite{shi2016end})
as the recognition head for scene (resp.
handwritten)
text.

\item 
In TREFE,
architectures 
not meeting the resource constraints
are dropped 
during search 
(Algorithm~\ref{alg:search_supernet}). 
Thus,
TREFE is deployment-aware. 
\end{enumerate}
As will be shown empirically in Sections~\ref{sec:varres} and~\ref{sec:exp:comp},
the above 
contribute to 
performance improvements over AutoSTR.




\section{Experiments}
\label{sec:exp}

In this section, 
we 
demonstrate the effectiveness of TREFE 
on long text (i.e., line level)~\cite{wang2020decoupled} and short text (i.e., word level)~\cite{wang2020decoupled}
recognition by
performing extensive experiments 
on handwritten and scene text TR datasets.


\subsection{Setup}
\label{sec:setup}


\noindent
\textbf{Handwritten TR Datasets.}
The following datasets are used:
\begin{enumerate}
[leftmargin=*]
\item \textbf{IAM}~\cite{MartiB02}: 
This dataset contains English handwritten text passages. 
The 
training set
contains 6482 lines from 747 documents.
The validation set
contains 976 lines from 116 documents,
and 
the test set contains
2915 lines from 336 documents.
The number of characters
is 80.
\item \textbf{RIMES}~\cite{grosicki2011icdar}: 
This contains French handwritten mail text 
from 1300 different writers.
The training set
contains 10532 lines from 1400 pages.
The validation
set
contains 801 lines from 100 pages,
and the test set contains
778 lines from 100 pages.
The number of characters is 100.
\end{enumerate}
The input image is of size 64$\times$1200. Moreover,
image augmentation is used as in~\cite{wang2020decoupled}.



\noindent
\textbf{Scene TR Datasets.}
Following~\cite{shi2018aster,yue2020robustscanner}, 
the task is to 
recognize
all 36 case-insensitive alphanumeric characters from the scene text.
Both synthetic datasets and real scene image datasets are used.
The input image is resized to 64$\times$256.  
The two synthetic datasets are:
\begin{enumerate}[leftmargin=*]
\item SynthText (\textbf{ST})\cite{liu2018synthetically}:
This contains 9 million text images 
generated from a lexicon of 90k common English words.
Words are rendered to images using 
various background and image transformations.

\item MJSynth (\textbf{MJ})\cite{jaderberg2014synthetic}:
This contains 6 million text images
cropped from 800,000 synthetic natural images with
ground-truth word bounding boxes.

\end{enumerate}
The 
real scene image datasets include:
\begin{enumerate}[leftmargin=*]
\item IIIT 5K-Words (\textbf{IIIT5K})~\cite{mishra2012top}:
This contains 5,000 cropped word images 
collected from the web.
2,000 images are used for training and the remaining 3,000 for testing.

\item Street View Text (\textbf{SVT})~\cite{wang2011end}:
This is harvested from Google Street View. 
The training set contains 257 word images, and
the test set contains 647 word images.
It exhibits high variability and the images often have low resolution. 

\item ICDAR 2003 (\textbf{IC03})~\cite{lucas2005icdar}:
This contains 251 full-scene text images. 
It contains 1,156 training images. 
Following~\cite{wang2011end}, we discard
test
images with non-alphanumeric characters or have fewer than three characters. 
As in~\cite{baek2019wrong},
two versions 
are used for testing:
one with 867 
images, and
the other has 860.

\item ICDAR 2013 (\textbf{IC13})~\cite{karatzas2013icdar}:
This contains 848 training images.
Two versions are used for testing:
one has 857 
images,
and the
other has 1,015.

\item ICDAR 2015 (\textbf{IC15}):
This is from the 4th Challenge in the ICDAR 2015 Robust Reading Competition~\cite{karatzas2015icdar}.
The data are collected via Google glasses without careful positioning and focusing. 
As a result, 
there are a lot of blurred 
images
in multiple orientations.
The training set
has 4468 
images.
Two versions of testing
datasets
are used: one has
1,811 
images
and the other has 2,077.

\item SVT-Perspective (\textbf{SVTP}):
This is used in~\cite{quy2013recognizing} for the evaluation
of 
perspective text 
recognition
performance.
Samples are selected from side-view images in Google Street View, and so
many of them are heavily deformed by perspective distortion.
It contains 645 test images. 
\end{enumerate}

As in~\cite{baek2019wrong}, we 
train the model using the two synthetic datasets.
The validation data is
formed 
by 
combining
the training sets of IC13, IC15, IIIT5K, and SVT.
The model is then evaluated on the test sets of the real scene image datasets
without fine-tuning.



\noindent
\textbf{Performance Evaluation.}
As in~\cite{wang2020decoupled,yousef2020origaminet}, the
following  measures are
used
for performance evaluation
on the handwritten text datasets:
\begin{enumerate}[leftmargin=*]
\item Character Error Rate (CER) 
$
=
\sum\nolimits_{i=1}^{G} \textit{edit}(y_i,\hat{y}_i)$ 
$/$ 
$\sum\nolimits_{j=1}^{G} \textit{length}(y_i)$,
where
$G$ is the dataset size,
$y_i$ is the ground-truth text, $\hat{y}_i$ is the predicted text, 
and $\textit{edit}$ is the Levenshtein distance~\cite{levenshtein1966binary};
\item Word Error Rate (WER):  defined in the same manner as CER, but at word level instead of character level.
\end{enumerate}
For the scene text datasets, 
following~\cite{shi2018aster,zhan2019esir,yang2019symmetry,wang2020decoupled},
we use
word accuracy for performance evaluation.
Moreover,
we also report the speed by measuring network latency on a NVIDIA 3090 GPU device
with the TensorRT library\footnote{\url{https://developer.nvidia.com/tensorrt}}.
Specifically,
following~\cite{chen2020fasterseg},
we use artificial
images as input
and perform
inference 
on the TR network
1000 times.
The average time is recorded as network latency.

\begin{table*}[ht]
	\caption{Comparison with the state-of-the-arts on the 
		scene text datasets.
		The number under the dataset name is the corresponding number of test samples.
		Word accuracies for the baselines are copied from the respective papers (`-'
		means that the corresponding result is not unavailable).
		The best result is in bold and 
		the second-best is underlined.}
	\label{tab:results_scene}
	\centering
	\vspace{-10px}
	\begin{tabular}{c|c|c|cc|cc|cc|c||c}
		\toprule
		                                          &                                                                   \multicolumn{9}{c||}{word accuracy}                                                                    &                              \\
		            \multirow{2}{*}{}             &      IIIT5K      &       SVT
						&      \multicolumn{2}{c|}{IC03}      &
						\multicolumn{2}{c|}{IC13}      &      \multicolumn{2}{c|}{IC15}
						&       SVTP       & \multirow{2}{*}{latency (ms)} \\
		                                          &       3000       &       647        &       860        &       867        &       857        &       1015       &       1811       &       2077       &       645        &                              \\ \midrule
		         AON~\cite{cheng2018aon}          &       87.0       &       82.8       &        -         &       91.5       &        -         &        -         &        -         &       68.2       &       73.0       &              -               \\
		          EP~\cite{bai2018edit}           &       88.3       &       87.5       &        -         &  \textbf{94.6}   &        -         & \underline{94.4} &        -         &       73.9       &        -         &              -               \\
		          SAR~\cite{li2019show}           &       91.5       &       84.5       &        -         &        -         &        -         &       91.0       &       69.2       &        -         &       76.4       &             4.58             \\
		        ESIR~\cite{zhan2019esir}          &       93.3       &       90.2       &        -         &        -         &       91.3       &        -         &        -         &       76.9       &       79.6       &              -               \\
		        ASTER~\cite{shi2018aster}         &       93.4       &       89.5       &       94.5       &        -         &        -         &       91.8       &        -         &       76.1       &       78.5       &             3.18             \\
		      SCRN~\cite{yang2019symmetry}        &       94.4       &       88.9       & \underline{95.0} &                  &        -         &       93.9       &        -         &  \textbf{80.8}   &       78.7       &              -               \\
		      DAN~\cite{wang2020decoupled}        &       93.3       &       88.4       &  \textbf{95.2}   &        -         &       94.2       &        -         &        -         &       71.8       &       76.8       &       \underline{2.92}       \\
		        SRN~\cite{yu2020towards}          & \underline{94.8} & \underline{91.5} &        -         &        -         &  \textbf{95.5}   &        -         & \underline{82.7} &        -         &  \textbf{85.1}   &             3.11             \\
		  TextScanner~\cite{wan2020textscanner}   &       93.9       &       90.1       &        -         &        -         &        -         &       92.9       &        -         &       79.4       &       83.7       &              -               \\
		RobustScanner~\cite{yue2020robustscanner} &  \textbf{95.3}   &       88.1       &        -         &        -         &        -         &  \textbf{94.8}   &        -         &       77.1       &       79.5       &             4.17             \\
		      PREN~\cite{yan2021primitive}        &       92.1       &  \textbf{92.0}   &       94.9       &        -         &       94.7       &        -         &        -         &       79.2       &       83.9       &             3.75             \\ 
		      		     AutoSTR~\cite{zhang2020autostr}      &       94.7       &       90.9       &       93.3       &        -         &       94.2       &        -         &       81.7       &        -         &       81.8       &              3.86               \\\midrule
		                  TREFE                   & \underline{94.8} &       91.3       &       93.7       & \underline{93.4} & \underline{95.4} &       93.0       &  \textbf{84.0}   & \underline{80.2} & \underline{84.5} &        \textbf{2.62}         \\ \bottomrule
	\end{tabular}
	\vspace{-10px}
\end{table*}



\noindent
\textbf{Implementation Details.}
For scene TR, 
a Spatial Transformer Network for rectification \cite{shi2018aster}
is used for pre-processing.
No extra pre-processing network is used for handwritten TR.
The TR head
is based on CTC~\cite{shi2016end}
for handwritten TR,
and parallel attention~\cite{yue2020robustscanner} 
for scene TR.
The proposed method is developed 
under the PyTorch framework.
We deploy models via TensorRT
for high-performance inference and 
measure the network latency 
with the FP32 (32-bit floating point computation) mode as in ~\cite{chen2020fasterseg}.

We use 20 MBConv layers for the spatial model and 
4 transformer layers for the sequential model.
To reduce computational complexity,
the supernet starts with a fixed ``stem'' layer that reduces the spatial resolution
with stride 2$\times$2.
In Algorithm~\ref{alg:prog},
we set $K=5$ and train each block for 300 epochs on the IAM training set.
For the larger scene text datasets, 
we train each block for 1 epoch.
Following~\cite{shi2018aster,yu2020towards},
we use ADADELTA~\cite{zeiler2012adadelta} 
with cosine learning rate 
as optimizer.
The initial learning rate is 0.8, 
and a weight decay of $10^{-5}$. 
The batch size 
is 64
for handwritten TR,
and 256
for scene TR.
The entire architecture search optimization 
takes about 3 days for handwritten TR,  and
5 days for scene TR.

Before evaluating the obtained architecture on a target dataset, 
we first retrain the whole TR system
from scratch.
Using the feature extractor architecture obtained
from the search procedure,
the TR system
is 
optimized 
by the ADADELTA optimizer with weight decay of $10^{-5}$. 
A cosine schedule is used
to anneal the learning rate from 0.8 to 0.
We train the network for 
1000 (resp. 6) epochs 
with a batch size of 64 (resp. 560)
for handwritten (resp. scene) TR.


\subsection{Comparison with the State-of-the-Arts}
\label{sec:exp:sota}

In this section, 
we compare TREFE with the state-of-the-art methods on handwritten text and scene
TR.
For simplicity, we do not use any lexicon or language model. 


\subsubsection{Scene Text Recognition}
\label{sec:exp:scene}

In this experiment, we compare  with the following state-of-the-arts:
(i) AON~\cite{cheng2018aon}, which extracts directional features to boost
recognition; (ii)
EP~\cite{bai2018edit}, which uses edit-distance-based sequence modeling;
(iii)
SAR~\cite{li2019show}, which introduces 2D attention; (iv)
ASTER~\cite{shi2018aster}, which uses a rectification network for irregular-sized
images; (v)
SCRN~\cite{yang2019symmetry}, which improves rectification with 
text shape description and explicit symmetric constraints;
(vi)
ESIR~\cite{zhan2019esir}, which iterates image rectification; (vii)
AutoSTR~\cite{zhang2020autostr},
which is the method proposed in the conference
version of this paper; (viii)
DAN~\cite{wang2020decoupled}, which decouples alignment attention with historical
decoding; (ix)
SRN~\cite{yu2020towards}, which uses a faster parallel decoding and semantic
reasoning block; (x)
TextScanner
\cite{wan2020textscanner}, which is based on segmentation
and uses  a mutual-supervision branch to more accurately locate the
characters; (xi)
RobustScanner~\cite{yue2020robustscanner}, which dynamically fuses a hybrid branch
and a position enhancement branch; (xii)
PREN~\cite{yan2021primitive}, which learns a primitive representation using pooling and weighted aggregator.
Both SCRN and
TextScanner
also use character box annotations.

Table~\ref{tab:results_scene}
shows the results. 
As can be seen,
TREFE has
comparable recognition performance and the lowest latency.
This demonstrates the effectiveness and efficiency of TREFE.

\subsubsection{Handwritten Text Recognition}
\label{sec:exp:handwr}

In this experiment, 
we compare TREFE with the following state-of-the-arts:
(i)
Bluche et al.~\cite{Bluche16}, which uses a deep architecture with
multidimensional LSTM 
to extract features for text
recognition; (ii)
Sueiras et al.~\cite{sueiras2018offline}, which extracts image patches and
then decodes characters via a sequence-to-sequence architecture with the addition
of a convolutional network; (iii)
Chowdhury et al.~\cite{chowdhury2018efficient}, which proposes an 
attention-based sequence-to-sequence network;
(iv)
Bhunia et al.~\cite{bhunia2019handwriting}, which uses an adversarial feature
deformation module that learns to elastically warp the extracted features; (v)
Zhang et al.~\cite{zhang2019sequence}, which uses a sequence-to-sequence
domain adaptation network to handle various handwriting styles; (vi)
Fogel et al.~\cite{fogel2020scrabblegan}, which generates handwritten text images
using a generative adversarial network (GAN); (vii)
Wang et al.~\cite{wang2020decoupled}, 
which 
alleviates the alignment problem in the attention mechanism of sequence-to-sequence text
recognition models; (viii)
Coquenet et al.~\cite{Coquenet2020},
which replaces the sequential model with lightweight, parallel convolutional
networks;
and (ix) Yousef et al.~\cite{yousef2020origaminet}, 
which does not use a sequential model but instead applies convolutional blocks 
with a gating mechanism;
(x) Shi et al.~\cite{shi2016end},
which uses VGG as the spatial model and BiLSTM as the sequential model,
and
(xi) AutoSTR~\cite{zhan2019esir}.
We do not compare with STR-NAS~\cite{hong2020memory} (which is concurrent with an
earlier conference version~\cite{zhang2020autostr} of TREFE) as its reported
performance is significantly worse.

\begin{table}[ht]
	\centering
	\caption{Comparison with the state-of-the-arts on the IAM dataset.  The best result is in bold and 
		the second-best is underlined.}
	\vspace{-10px}
	\begin{tabular}{c|c|c|c}
		\toprule
		                                               &     WER (\%)      &
																	  CER (\%)     &   latency(ms)    \\ \midrule
		        Bluche et al.~\cite{Bluche16}          &       24.60       &       7.90       &        -         \\ 
		   Sueiras et al.~\cite{sueiras2018offline}    &       23.80       &       8.80       &        -         \\ 
		Chowdhury et al.~\cite{chowdhury2018efficient} & \underline{16.70} &       8.10       &       8.71       \\ 
		  Bhunia et al.~\cite{bhunia2019handwriting}   &       17.19       &       8.41       &        -         \\ 
		    Zhang et al.~\cite{zhang2019sequence}      &       22.20       &       8.50       &      10.52       \\ 
		   Fogel et al.~\cite{fogel2020scrabblegan}    &       23.61       &        -         &      12.16       \\ 
		     Wang et al.~\cite{wang2020decoupled}      &       20.60       &       7.00       &       7.64       \\ 
		     Coquenet et al.~\cite{Coquenet2020}       &       28.61       &       7.99       &  \textbf{2.08}   \\ 
		  Yousef et al.~\cite{yousef2020origaminet}    &         -         & \underline{4.76} &      21.48       \\ 
		         Shi et al.~\cite{shi2016end}          &       21.67       &       6.28       &       4.71       \\
		       AutoSTR~\cite{zhang2020autostr}         &       45.23       &      26.24       &      11.42       \\ \midrule
		TREFE          &  \textbf{16.41}   &  \textbf{4.45}   & \underline{2.85} \\ \bottomrule
	\end{tabular}
	\label{tab:results_iam}
\end{table}

Tables~\ref{tab:results_iam}
and~\ref{tab:results_rimes}
show results on the IAM and RIMES datasets, respectively.
For the baselines,
their WER's and CER's 
are copied from the respective
papers\footnote{Note that Yousef et al.~\cite{yousef2020origaminet}      does not
report the WER.}, while
their latencies
are obtained by measurements on our reimplementations.\footnote{We do not report latency results for Bluche et al.~\cite{Bluche16}, Sueiras et al.~\cite{sueiras2018offline} and Bhunia et al.~\cite{bhunia2019handwriting}, as some implementation details are missing.}
Note that these two datasets have different number 
of characters,
\footnote{Methods in Zhang et al.~\cite{zhang2019sequence}
	and Yousef et al.~\cite{yousef2020origaminet}  do not have results 
	on RIMES, while Wang et al.~\cite{wang2020decoupled}
	only report
	results on an ensemble.}
thus the latency for IAM and RIMES are different.
As can be seen, 
AutoSTR cannot obtain good architecture 
and its latency is large.
The
architecture 
obtained  by TREFE 
has the best WER and CER performance. 
While
the method in Coquenet et al.~\cite{Coquenet2020} has the lowest latency,
the TREFE model has much lower error rates.


\begin{table}[ht]
	\centering
\caption{Comparison with the state-of-the-arts on the RIMES dataset. 
The best result is in bold and 
		the second-best is underlined.}
	\vspace{-10px}
	\begin{tabular}{c|c|c|c}
		\toprule
		& WER(\%)          & CER(\%)          &  latency(ms)          \\  \midrule
		Bluche et al.~\cite{Bluche16}			       & 12.60            & \underline{2.90} & -                     \\  
		Sueiras et al.~\cite{sueiras2018offline}       & 15.90            & 4.80             & -                     \\  
		Chowdhury et al.~\cite{chowdhury2018efficient} & \underline{9.60} & 3.50             &  9.11                 \\ 
		Bhunia et al.~\cite{bhunia2019handwriting}     & 10.47            & 6.44             & -                     \\ 
		Fogel et al.~\cite{fogel2020scrabblegan}       & 11.32            &  -               &  12.38                \\
		Coquenet et al.~\cite{Coquenet2020}            & 18.01            & 4.35             & \textbf{2.09}         \\
		Shi et al.~\cite{shi2016end}			       & 11.15		      & 3.40		     & 4.73 			     \\   
		AutoSTR~\cite{zhang2020autostr}                & 20.40            & 11.31            & 12.31 \\ \midrule 
		TREFE										   & \textbf{9.16}    & \textbf{2.75}    & \underline{2.86}      \\
		\bottomrule
\end{tabular}
\label{tab:results_rimes}
\end{table}

\begin{table*}[ht]
	\centering
	\caption{Architectures obtained on the IAM dataset (left) and scene text dataset (right).
		$\checkmark$ (\textit{resp.} $\times$) means to follow alternative (\textit{resp.} original) choices in Section~\ref{sec:rnn_search_space}.}
	\setlength{\tabcolsep}{1mm}
	\vspace{-10px}
	
	\begin{tabular}{c|c|c|c|c|c|c}
		\toprule
		\multirow{22}{*}{$\M{N}_\text{spa}^*$} & layer & \multicolumn{2}{c|}{operator} &
		\multicolumn{2}{c|}{resolution} & \#channels  \\  \cmidrule{2-7} 
		& stem    & \multicolumn{2}{c|}{Conv(k:3)-BN-ReLU}           & \multicolumn{2}{c|}{[32, 600]} & 16       \\  
		& 1       & \multicolumn{2}{c|}{MBConv(k:5,e:6)  }           & \multicolumn{2}{c|}{[32, 600]} & 16       \\  
		& 2       & \multicolumn{2}{c|}{MBConv(k:5,e:6)  }           & \multicolumn{2}{c|}{[32, 600]} & 16       \\  
		& 3       & \multicolumn{2}{c|}{MBConv(k:5,e:6)  }           & \multicolumn{2}{c|}{[16, 600]} & 16       \\  
		& 4       & \multicolumn{2}{c|}{MBConv(k:5,e:6)  }           & \multicolumn{2}{c|}{[16, 600]} & 16       \\  
		& 5       & \multicolumn{2}{c|}{MBConv(k:5,e:6)  }           & \multicolumn{2}{c|}{[16, 600]} & 16       \\  
		& 6       & \multicolumn{2}{c|}{MBConv(k:5,e:6)  }           & \multicolumn{2}{c|}{[16, 600]} & 16       \\  
		& 7       & \multicolumn{2}{c|}{MBConv(k:5,e:6)  }           & \multicolumn{2}{c|}{[16, 600]} & 16       \\  
		& 8       & \multicolumn{2}{c|}{MBConv(k:3,e:1)  }           & \multicolumn{2}{c|}{[16, 600]} & 16       \\  
		& 9       & \multicolumn{2}{c|}{MBConv(k:5,e:6)  }           & \multicolumn{2}{c|}{[8, 600]}  & 32       \\  
		& 10      & \multicolumn{2}{c|}{MBConv(k:5,e:6)  }           & \multicolumn{2}{c|}{[8, 600]}  & 32       \\  
		& 11      & \multicolumn{2}{c|}{MBConv(k:5,e:6)  }           & \multicolumn{2}{c|}{[8, 600]}  & 32       \\  
		& 12      & \multicolumn{2}{c|}{MBConv(k:5,e:6)  }           & \multicolumn{2}{c|}{[8, 600]}  & 32       \\  
		& 13      & \multicolumn{2}{c|}{MBConv(k:5,e:1)  }           & \multicolumn{2}{c|}{[4, 600]}  & 64       \\  
		& 14      & \multicolumn{2}{c|}{MBConv(k:5,e:1)  }           & \multicolumn{2}{c|}{[2, 300]}  & 128      \\  
		& 15      & \multicolumn{2}{c|}{MBConv(k:5,e:1)  }           & \multicolumn{2}{c|}{[2, 300]}  & 128      \\  
		& 16      & \multicolumn{2}{c|}{MBConv(k:5,e:6)  }           & \multicolumn{2}{c|}{[2, 300]}  & 128      \\  
		& 17      & \multicolumn{2}{c|}{MBConv(k:5,e:1)  }           & \multicolumn{2}{c|}{[2, 300]}  & 128      \\  
		& 18      & \multicolumn{2}{c|}{MBConv(k:5,e:1)  }           & \multicolumn{2}{c|}{[2, 300]}  & 128      \\  
		& 19      & \multicolumn{2}{c|}{MBConv(k:5,e:6)  }           & \multicolumn{2}{c|}{[1, 150]}  & 256      \\  
		& 20      & \multicolumn{2}{c|}{MBConv(k:5,e:6)  }           & \multicolumn{2}{c|}{[1, 150]}  & 256      \\  \midrule 
		\multirow{5}{*}{$\M{N}_\text{seq}^*$} & layer      & (i):residual         & (ii):Rel       &
		(iii):scaling       & (iv):FFN         & \#hidden      \\  \cmidrule{2-7}
		& 21      		  & $\times$     & $\times$     & $\checkmark$   & GLU         & 256   \\  
		& 22      		  & $\checkmark$ & $\times$     & $\checkmark$   & MLP         & 256   \\  
		& 23      		  & $\times$     & $\checkmark$ & $\times$       & GLU         & 256   \\  
		& 24      		  & $\checkmark$ & $\checkmark$ & $\checkmark$   & GLU         & 256   \\  
		\bottomrule
	\end{tabular}
	\qquad
	\begin{tabular}{c|c|c|c|c|c|c}
		\toprule
		\multirow{22}{*}{$\M{N}_\text{spa}^*$} & layer &
		\multicolumn{2}{c|}{operator} & \multicolumn{2}{c|}{resolution} & \#channels  \\  \cmidrule{2-7} 
		& stem    & \multicolumn{2}{c|}{Conv(k:3)-BN-ReLU}           & \multicolumn{2}{c|}{[32, 128]}  & 32       \\ 
		& 1       & \multicolumn{2}{c|}{MBConv(k:5,e:6)  }           & \multicolumn{2}{c|}{[32, 128]}  & 32       \\  
		& 2       & \multicolumn{2}{c|}{MBConv(k:3,e:6)  }           & \multicolumn{2}{c|}{[16, 128]}  & 32       \\  
		& 3       & \multicolumn{2}{c|}{MBConv(k:5,e:6)  }           & \multicolumn{2}{c|}{[16, 128]}  & 32       \\  
		& 4       & \multicolumn{2}{c|}{MBConv(k:5,e:6)  }           & \multicolumn{2}{c|}{[16, 128]}  & 32       \\  
		& 5       & \multicolumn{2}{c|}{MBConv(k:3,e:6)  }           & \multicolumn{2}{c|}{[16, 128]}  & 32       \\  
		& 6       & \multicolumn{2}{c|}{MBConv(k:3,e:1)  }           & \multicolumn{2}{c|}{[16, 128]}  & 32       \\  
		& 7       & \multicolumn{2}{c|}{MBConv(k:5,e:6)  }           & \multicolumn{2}{c|}{[16, 128]}  & 32       \\  
		& 8       & \multicolumn{2}{c|}{MBConv(k:5,e:1)  }           & \multicolumn{2}{c|}{[16, 128]}  & 32       \\  
		& 9       & \multicolumn{2}{c|}{MBConv(k:5,e:1)  }           & \multicolumn{2}{c|}{[16, 128]}  & 32       \\  
		& 10      & \multicolumn{2}{c|}{MBConv(k:5,e:6)  }           & \multicolumn{2}{c|}{[16, 128]}  & 32       \\  
		& 11      & \multicolumn{2}{c|}{MBConv(k:3,e:6)  }           & \multicolumn{2}{c|}{[8, 64]}    & 64       \\  
		& 12      & \multicolumn{2}{c|}{MBConv(k:3,e:6)  }           & \multicolumn{2}{c|}{[8, 64]}    & 64       \\  
		& 13      & \multicolumn{2}{c|}{MBConv(k:3,e:6)  }           & \multicolumn{2}{c|}{[4, 64]}    & 128       \\  
		& 14      & \multicolumn{2}{c|}{MBConv(k:5,e:6)  }           & \multicolumn{2}{c|}{[4, 64]}    & 128      \\  
		& 15      & \multicolumn{2}{c|}{MBConv(k:5,e:6)  }           & \multicolumn{2}{c|}{[4, 64]}    & 128      \\  
		& 16      & \multicolumn{2}{c|}{MBConv(k:3,e:6)  }           & \multicolumn{2}{c|}{[4, 64]}    & 128      \\  
		& 17      & \multicolumn{2}{c|}{MBConv(k:3,e:6)  }           & \multicolumn{2}{c|}{[4, 64]}    & 128      \\  
		& 18      & \multicolumn{2}{c|}{MBConv(k:5,e:6)  }           & \multicolumn{2}{c|}{[2, 64]}    & 256      \\  
		& 19      & \multicolumn{2}{c|}{MBConv(k:5,e:6)  }           & \multicolumn{2}{c|}{[1, 32]}    & 512      \\  
		& 20      & \multicolumn{2}{c|}{MBConv(k:5,e:6)  }           & \multicolumn{2}{c|}{[1, 32]}    & 512      \\  \midrule
		\multirow{5}{*}{$\M{N}_\text{seq}^*$} & layer      & (i):residual         & (ii):Rel       &
		(iii):scaling       & (iv):FFN         & \#hidden      \\  \cmidrule{2-7}
		& 21      & $\times$     & $\checkmark$     & $\checkmark$   & MLP         & 512   \\  
		& 22      & $\times$     & $\checkmark$     & $\checkmark$   & MLP         & 512   \\  
		& 23      & $\checkmark$ & $\times$         & $\checkmark$   & MLP         & 512   \\  
		& 24      & $\checkmark$ & $\checkmark$     & $\checkmark$   & GLU         & 512   \\  \bottomrule
	\end{tabular}
	\label{tab:searched_archs}
\end{table*}


\subsection{Understanding Architectures Obtained by TREFE}
\label{sec:aba1}

In this section,
we provide a closer look at the architectures
obtained by the proposed TREFE.


\subsubsection{Feature Extractors}

Table~\ref{tab:searched_archs}
(left)
shows the 
spatial 
and sequential model
architectures 
($\M{N}_\text{spa}^*$ and $\M{N}_\text{seq}^*$, respectively)
obtained by TREFE 
on the IAM dataset.
For the first half of $\M{N}_\text{spa}^*$
(layers 1 to 10),
the resolution 
is relatively high 
and is only
slowly reduced by a total factor of 4. 
For the second half of $\M{N}_\text{spa}^*$
(layers 10 to 20),
the resolution 
is reduced more rapidly by a total factor of 32.
A similar observation can also be made on the architecture obtained on
the scene text dataset 
(Table~\ref{tab:searched_archs} (right)).
We speculate that a larger resolution can help the network 
to preserve more spatial information.
This is also consistent with 
some manually-designed network architectures.
For example, in SCRN~\cite{yang2019symmetry},
a ResNet50 with FPN~\cite{lin2017feature} 
is first used to extract features from text images. 
This is followed by
a few convolutional layers which 
downsample the feature map resolution rapidly by a factor of $32$.
This observation may inspire 
better designs of the text image feature extractor in the future.


\subsubsection{Varying the Resource Constraints}
\label{sec:varres}

In this section,
we demonstrate the deployment-aware ability of TREFE
by performing experiments with different latency constraints on the IAM dataset.
These include:
(i) no latency constraint,
(ii) reduce the runtime to $5/6$ of that of the architecture obtained under no latency constraint, and,
(iii) reduce the runtime  to $2/3$ of that of the architecture obtained under no latency constraint.
We compare TREFE with random search,
which is often a
strong baseline in NAS~\cite{li2019random}.
Specifically, 
we randomly pick $6$ architectures from the search space
that satisfy the required latency constraint, 
and then
train them from scratch 
using the same setting as in Section~\ref{sec:setup}.

Table~\ref{tab:resource_constrain}
shows the performance of models obtained by TREFE and random search.
As can be seen,
TREFE obtains architectures with better WER and CER.
Though the TREFE models have higher latencies,
the gap with random search closes rapidly as 
the target latency  is reduced.

\begin{table*}[ht]
	\centering
	\caption{Comparison of TREFE and random search under different latency
		constraints on the IAM dataset.}
	\vspace{-10px}
	\begin{tabular}{c|c|c|c|c|c|c}
		\toprule
		\multirow{2}{*}{Latency constraint} & \multicolumn{3}{c|}{TREFE } &
		\multicolumn{3}{c}{random search}  \\ 
		& WER (\%)   & CER (\%)    & latency (ms) & WER (\%)  & CER (\%)  & latency (ms) \\ \midrule
		(i) no constraint & \textbf{16.41} & \textbf{4.45}  & 2.85  & 17.04 & 4.68 &   \textbf{2.20}           \\ 
		(ii) reduce latency to 
		$\nicefrac{5}{6}$ of unconstrained model
		& \textbf{16.67} & \textbf{4.56} & 2.37 & 17.04 & 4.68 &   \textbf{2.20}  \\
		(iii) reduce latency to $\nicefrac{2}{3}$ of unconstrained model
		& \textbf{19.18} & \textbf{5.19} & 1.86 & 19.39 & 5.40 & \textbf{1.84}  \\
		\bottomrule
	\end{tabular}
	\label{tab:resource_constrain}
\end{table*}


\subsubsection{Varying the Spatial Model Design}
\label{sec:vary-spatial}

Here,
we compare
the 
spatial model
$\mathcal{N}_{\text{spa}}^*$
obtained  by TREFE
with the popular hand-designed architectures
of VGG~\cite{simonyan2014very} 
(used in CRNN~\cite{shi2016end})
and 
ResNet~\cite{he2016deep}
(used in ASTER~\cite{shi2018aster}).
We also compare 
with the spatial model 
obtained by DARTS~\cite{DARTS}, a
representative NAS method.
To ensure that 
the constraint in \eqref{eq:opt} is satisfied by the
DARTS
model,
we replace the basic block in ASTER
with a CNN cell from DARTS,
which is searched on the image classification task
using the ImageNet dataset.
All architectures are trained and evaluated 
under the same settings as in Section~\ref{sec:setup}.
The experiment is performed on the IAM, RIMES and IIIT5K datasets.
Note that random search is not compared,
since 
it has been shown to be less effective than 
TREFE in Section~\ref{sec:varres}.

Table~\ref{tab:improved_spa_model} shows the results.
As can be seen, 
TREFE
(i.e.,
$\mathcal{N}_{\text{spa}}^*$ + $\mathcal{N}_{\text{seq}}^*$)
outperforms
the other baselines on all datasets.
The performance of DARTS
is even worse than the hand-designed architectures.
This demonstrates that
ignoring the domain knowledge and
directly reusing structures obtained by a state-of-the-art NAS method 
may not be good.

\begin{table*}[ht]
	\centering
\caption{Performance comparison of different spatial model architectures.
		``VT'' is short for Vanilla Transformer,
		``DARTS'' means reusing CNN architecture searched by DARTS~\cite{DARTS}.}
	\vspace{-10px}
\begin{tabular}{c| cc|| c|c|c || c|c|c || c | c}
		\toprule
		&                       \multicolumn{2}{c||}{architecture}                       &            \multicolumn{3}{c||}{IAM}            &           \multicolumn{3}{c||}{RIMES}         &           \multicolumn{2}{c}{IIIT5K}  \\
		& spatial & sequential                                           & WER (\%)
		&   CER(\%)    & latency(ms)  &   WER(\%)    &   CER(\%)    & latency(ms) & Acc(\%) &  latency(ms) \\ \midrule
		hand-         & ResNet                       & VT                                            & 22.10          &     6.19      &     2.00      &     12.90     &     3.78      &     2.01      &   93.8   &  2.01   \\ \cmidrule{2-11}
		designed       & VGG                          & \multirow{2}{*}{$\mathcal{N}_{\text{seq}}^*$} & 20.80          &     5.97      & \textbf{1.21} &     11.47     &     3.38      & \textbf{1.21} &   92.8   &  1.27   \\
		& ResNet                       &                                               & 18.67          &     5.24      &     2.19      &     10.24     &     3.10      &     2.21      &   93.4   &  2.16   \\ \midrule
		\multirow{2}{*}{NAS} & DARTS                        & \multirow{2}{*}{$\mathcal{N}_{\text{seq}}^*$} & 23.24          &     7.02      &     3.81      &     12.03     &     3.81      &     3.92      &   91.9   &  3.19   \\
		& $\mathcal{N}_{\text{spa}}^*$ &                                               & \textbf{16.41} & \textbf{4.45} &     2.85      & \textbf{9.16} & \textbf{2.75} &     2.86      &   \textbf{94.8}   &  \textbf{2.62}   \\ \bottomrule
	\end{tabular}  
	\label{tab:improved_spa_model}
\end{table*}


\subsubsection{Varying the Sequential Model Design}
\label{sec:aba3}

In this experiment,
we compare 
the 
sequential model $\mathcal{N}_{\text{seq}}^*$
obtained  by TREFE
with the  (i) hand-designed BiLSTM in~\cite{hochreiter1997long}, 
which is a strong baseline for sequential context modeling in current text
recognition systems~\cite{shi2018aster,wang2020decoupled,wan2020textscanner},
and (ii) vanilla Transformer~\cite{VaswaniSPUJGKP17}.
We also compare with the sequential models obtained by two NAS methods:
(i) the recurrent cell obtained by DARTS~\cite{DARTS}
on the Penn Treebank; 
and (ii) the evolved Transformer~\cite{So2019TheET}, which is obtrained
on the WMT'14 En-De translation task~\cite{bojar2014findings}.
In these baselines,
we keep the spatial model $\mathcal{N}_{\text{spa}}^*$ in TREFE, but
replace its sequential model $\mathcal{N}_{\text{seq}}^*$
with the models
in these baselines.
The resultant architectures are trained (from scratch)
and evaluated 
under the same settings as in Section~\ref{sec:setup}.
The experiment is again performed on the IAM, RIMES and
IIIT5K 
datasets.

Table~\ref{tab:improved_seq_model} shows the results.
As can be seen, 
TREFE
(i.e.,
$\mathcal{N}_{\text{spa}}^*$ + $\mathcal{N}_{\text{seq}}^*$)
consistently
outperforms all 
the baselines, 
including the transformer (VT).
Thus, 
architecture search 
of the sequential model 
is necessary.

\begin{table*}[ht]
	\centering
	\caption{Performance comparison of different sequential model architectures.
		``VT'' is short for Vanilla Transformer,
		``DARTS'' means reusing RNN architecture searched by DARTS~\cite{DARTS},
		and ``ET'' is the Evolving transformer~\cite{So2019TheET}.}
	\vspace{-10px}
\begin{tabular}{c| cc|| c|c|c || c|c|c || c|c}
		\toprule
		&                      \multicolumn{2}{c||}{architecture}                       &            \multicolumn{3}{c||}{IAM}            &           \multicolumn{3}{c||}{RIMES}          & \multicolumn{2}{c}{IIIT5K} \\
		& spatial & sequential                          & WER(\%)       &   CER(\%)
		& latency(ms)  &   WER(\%)    &   CER(\%)    & latency(ms) & Acc(\%) &
		latency(ms) \\ \midrule
		hand-                   & ResNet                                        & VT                           & 22.10          &     6.19      &     2.00      &     12.90     &     3.78      &     2.01      &   93.8   &  2.01         \\	\cmidrule{2-11}
		designed                   & \multirow{2}{*}{$\mathcal{N}_{\text{spa}}^*$} & BiLSTM                       & 18.65          &     5.01      &     4.03      &     10.47     &     3.20      &     4.04      &   94.4   &  3.26         \\
		&                                               & VT                           & 19.84          &     5.29      & \textbf{2.64} &     11.79     &     3.41      & \textbf{2.65} &   94.3   &  2.45         \\ \midrule
		& \multirow{3}{*}{$\mathcal{N}_{\text{spa}}^*$} & DARTS                        & 19.42          &     5.28      &     28.35     &     10.97     &     3.23      &     28.86     &   92.8   &  8.57      \\
		NAS                    &                                               & ET                           & 20.27          &     5.56      &     3.27      &     11.77     &     3.33      &     3.27      &   93.1   &  3.18      \\
		&                                               & $\mathcal{N}_{\text{seq}}^*$ & \textbf{16.41} & \textbf{4.45} &     2.85      & \textbf{9.16} & \textbf{2.75} &     2.86      &   \textbf{94.8}   &  2.62      \\ \bottomrule
	\end{tabular}
	\label{tab:improved_seq_model}
\end{table*}


\subsection{Understanding the Search Process}
\label{sec:exp:proce}


\subsubsection{Supernet Training}
\label{sec:exp:spos}

In this section, we compare the 
proposed supernet training strategy (in Section~\ref{ssec:train_supernet}) with
the following strategies:
\begin{enumerate}[leftmargin=*]
\item SPOS~\cite{guo2020single}:
a widely-used one-shot NAS method 
which uniformly samples and updates $\bm{\alpha}$ 
from the supernet; 

\item  
Random path:
The proposed progressive supernet training pipeline 
(Algorithm~\ref{alg:prog}),
which randomly selects $\bm{\alpha}_k^{\leftarrow}$ and $\bm{\alpha}_k$; 

\item  
Best path: This is based on the proposed procedure,  but
instead of random sampling a path $\bm{\alpha}_k^{\leftarrow}$ from  the
trained $\Phi_1,\dots, \Phi_{k - 1}$,
it picks the $\bm{\alpha}_k^{\leftarrow}$
with the best validation performance 
(details are in Appendix~\ref{app:best_path});

\item  
Co-update:  This is also 
based on the proposed procedure,  but
instead of fixing the weights of $\bm{\alpha}_k^{\leftarrow}$, 
it updates them
together with 
$\mathbf{W}_k(\bm{\alpha}_k)$.
\end{enumerate}
For SPOS,
we train the whole supernet for 1500 epochs. 
To have a fair comparison,
for TREFE and its variations
((2), (3) and (4)), 
we first divide the supernet
into 5 blocks and 
then train each block for 300 epochs.
The total number of training epochs for all methods are thus the same.

\begin{figure*}[ht]
	\centering
	\subfigure[SPOS.]{\includegraphics[width=0.24\textwidth]{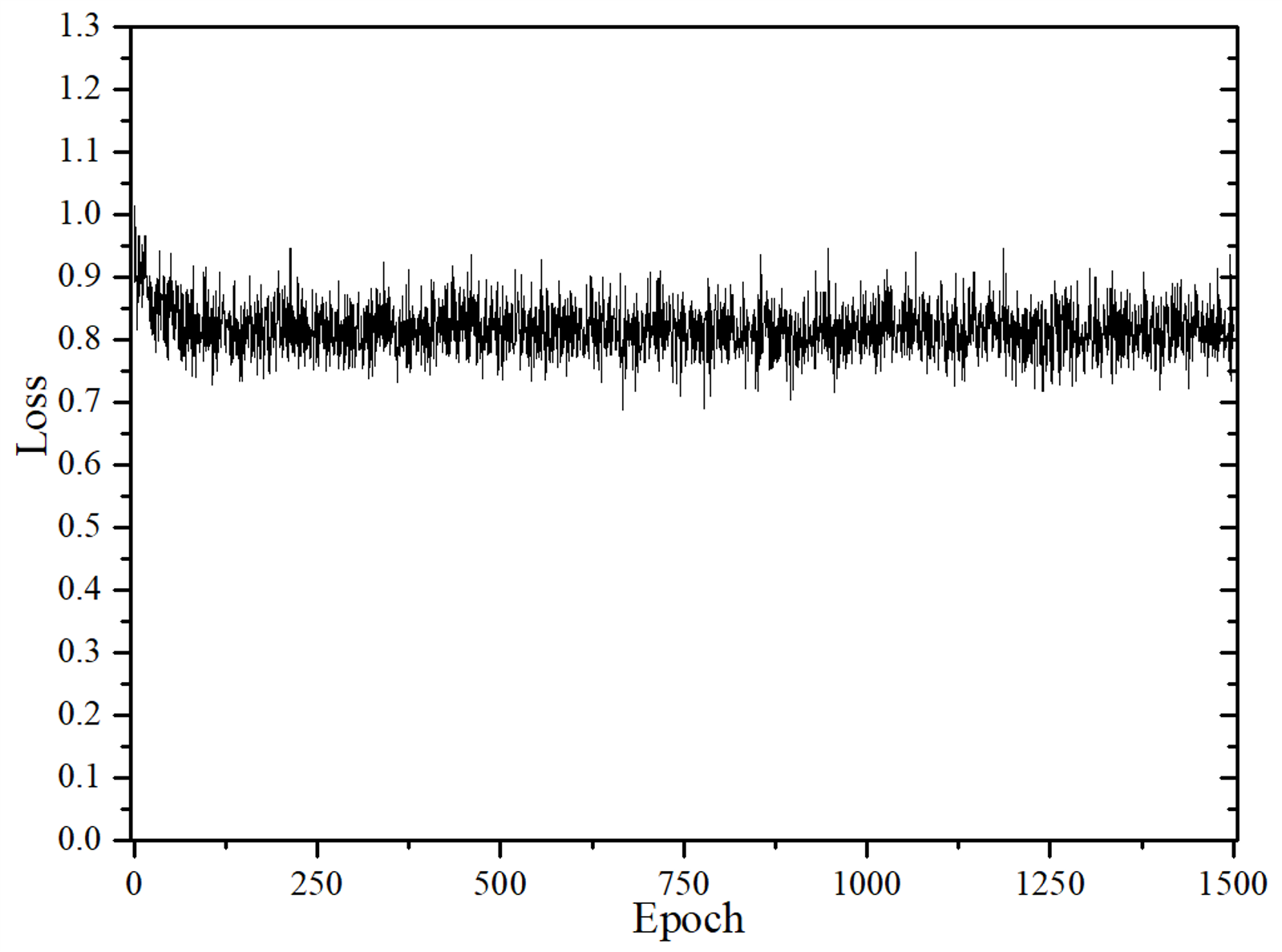}}
	\subfigure[Random path.]{\includegraphics[width=0.24\textwidth]{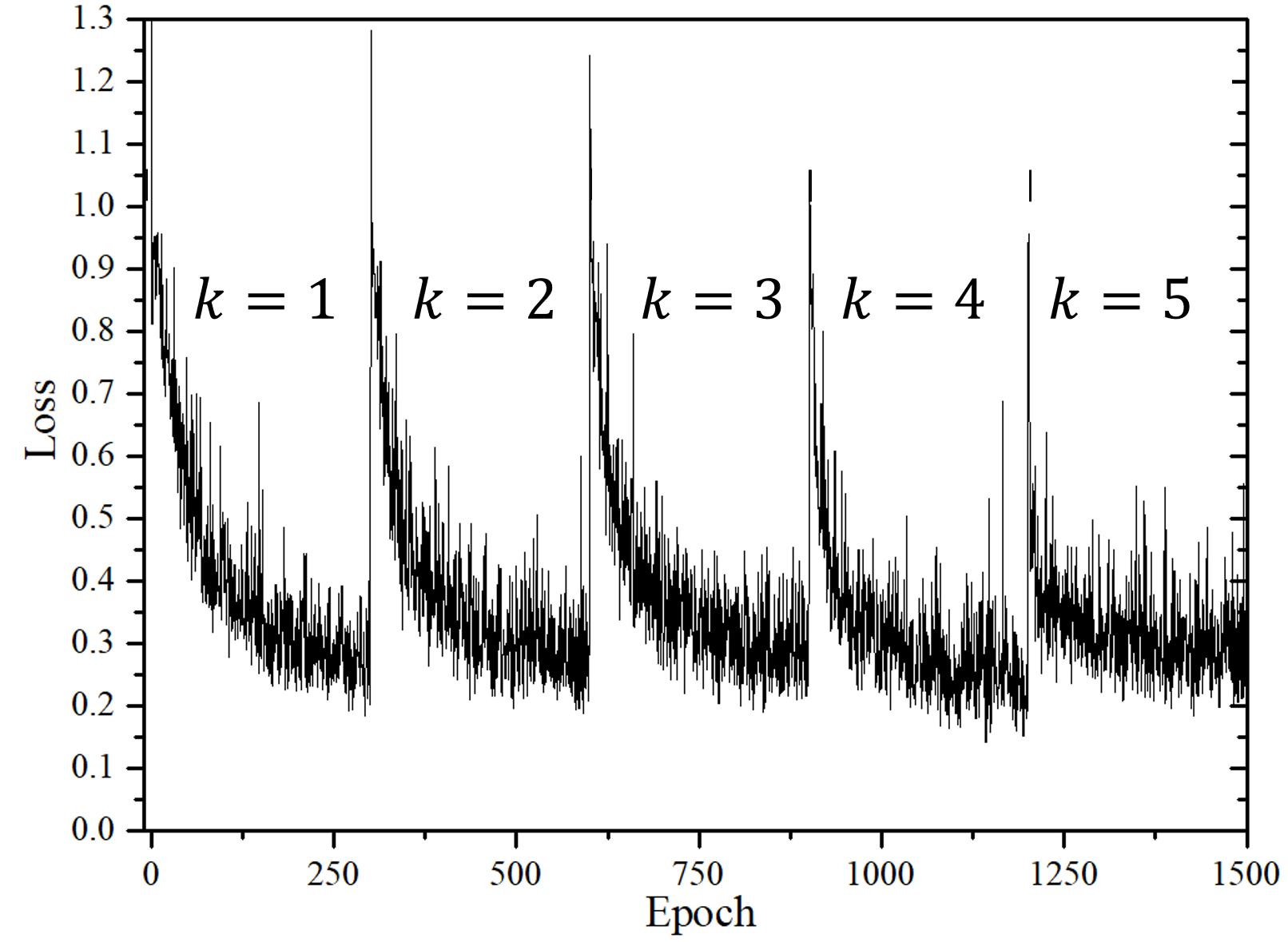}}
	\subfigure[Best path.]{\includegraphics[width=0.24\textwidth]{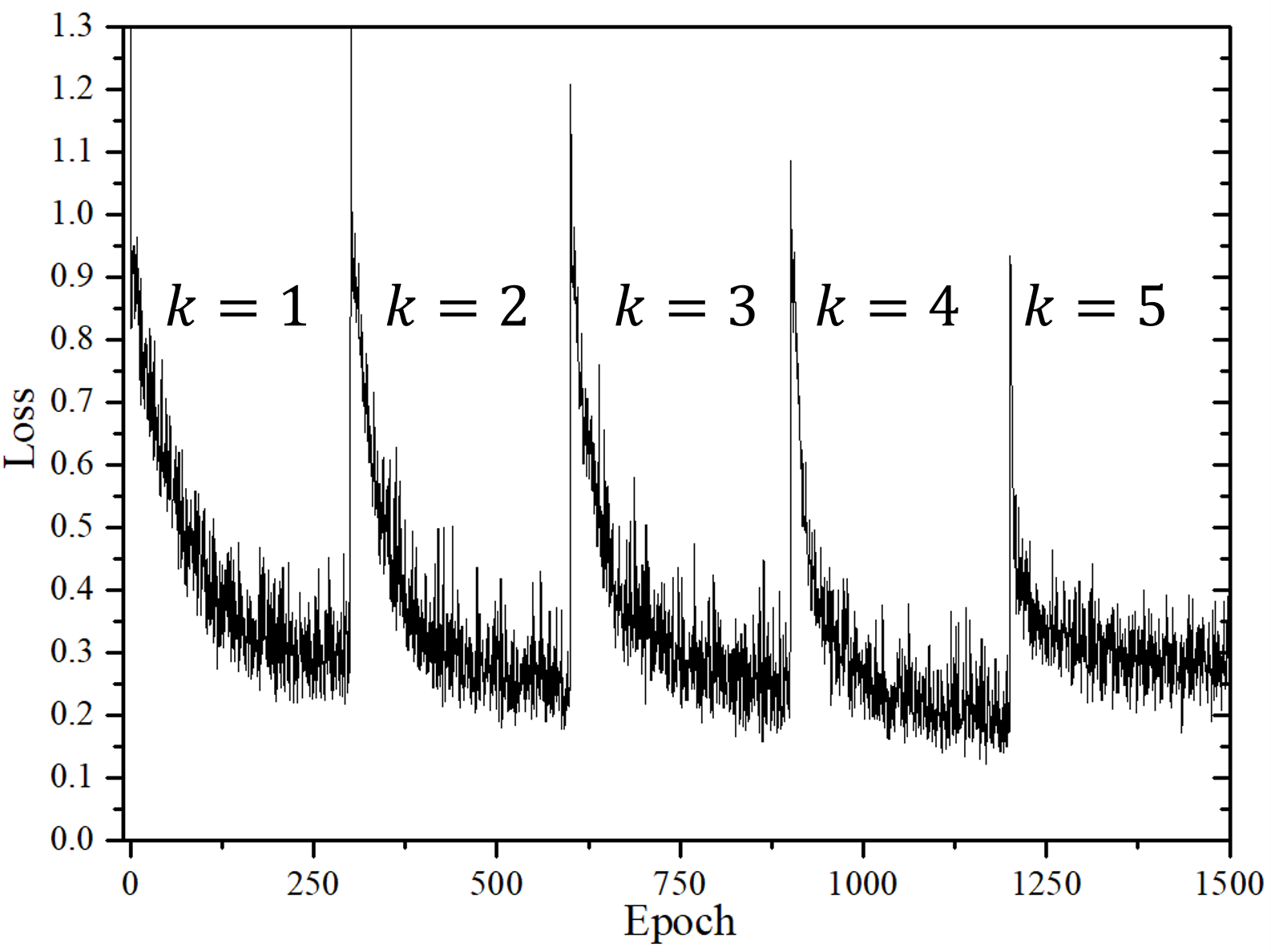}}
	\subfigure[Co-update.]{\includegraphics[width=0.24\textwidth]{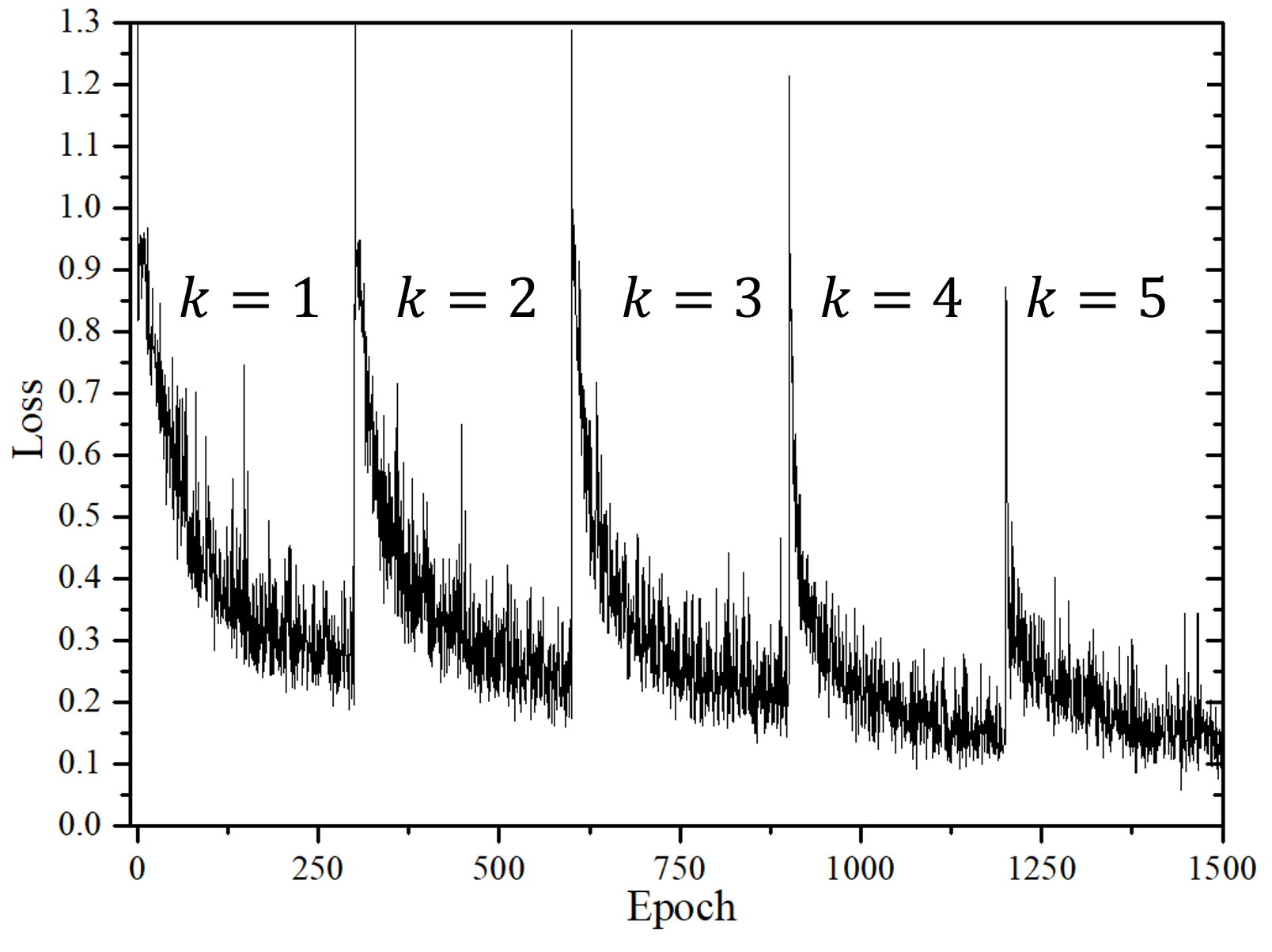}}
	
	\vspace{-10px}
	\caption{Comparison of the supernet training loss curves for
		SPOS~\cite{guo2020single}, TREFE (with random path) and its variations (best path and co-update).}
	\label{fig:supernet_training1}
\end{figure*}

\noindent
\textbf{Training.}
Figure~\ref{fig:supernet_training1}
shows the training losses for the various supernet training strategies
on the IAM dataset.
As can be seen, 
SPOS is difficult to converge, while TREFE and its variants show good training
convergence.

Table~\ref{tab:supernet} shows the 
training cost of the supernet.
As can be seen, ``random path'' and ``best path''
have lower training time than ``SPOS'' and ``co-update'',
as only parts of the selected path $\bm{\alpha}_k$ need to be updated.

\begin{table}[H]
\caption{Training costs (in GPU days) of the supernet.}
\centering
\vspace{-10px}
\setlength\tabcolsep{4pt}
\begin{tabular}{c|c|c|c} \toprule
SPOS &        Random path  & Best path   & Co-update   \\ \midrule
4.0  & 1.4 & 1.2 & 3.0 \\ \bottomrule
	\end{tabular}
	\label{tab:supernet}
\end{table}

\noindent
\textbf{Ranking Correlation.}
As in~\cite{li2020blockwisely,guo2020single},
we examine
the ranking correlation between the performance of a model trained from scratch
(denoted ``stand-alone model") and 
that with 
weights
inherited 
from the supernet ("one-shot model").
Specifically, 
for the stand-alone models,
we random sample 70 architectures from the proposed search space and 
train them from scratch for 250 epochs. 
The weights 
of the corresponding one-shot models
are obtained from the trained supernet and then 
finetuned
on the training set for 5 epochs (with learning rate 0.01).
As in~\cite{li2021bossnas}, 
the ranking correlation between the 
stand-alone model's
validation set CER
and that of
the one-shot model
is measured by the 
following
three commonly used metrics:
(i) Kendall's,
(ii) Spearman's $\rho$,
and (iii) Pearson's $r$.
They all have with values in $[-1, 1]$.

\begin{figure*}[ht]
	\centering
	\subfigure[SPOS.]{\includegraphics[width=0.24\textwidth]{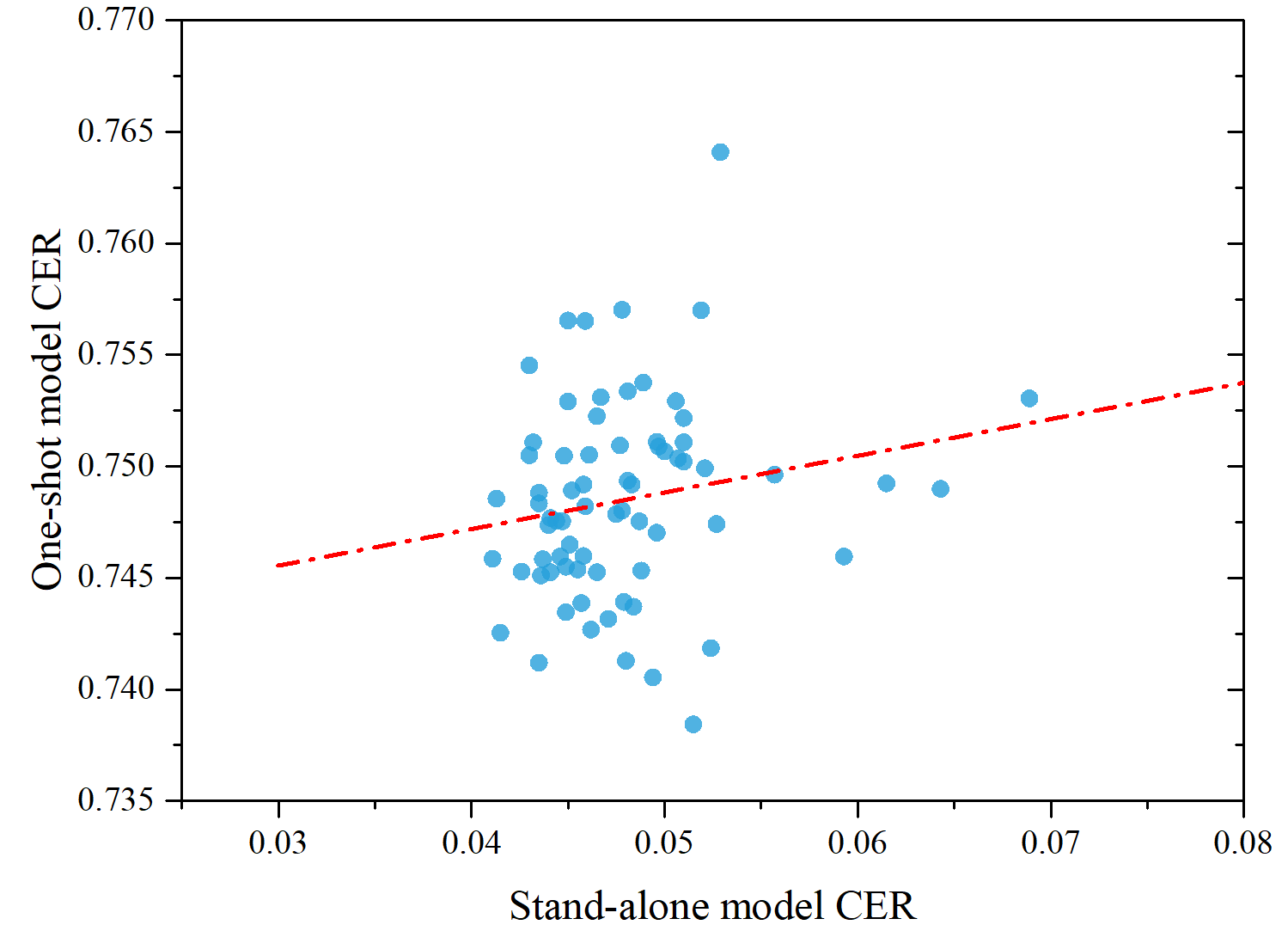}}
	\subfigure[Random path.]{\includegraphics[width=0.24\textwidth]{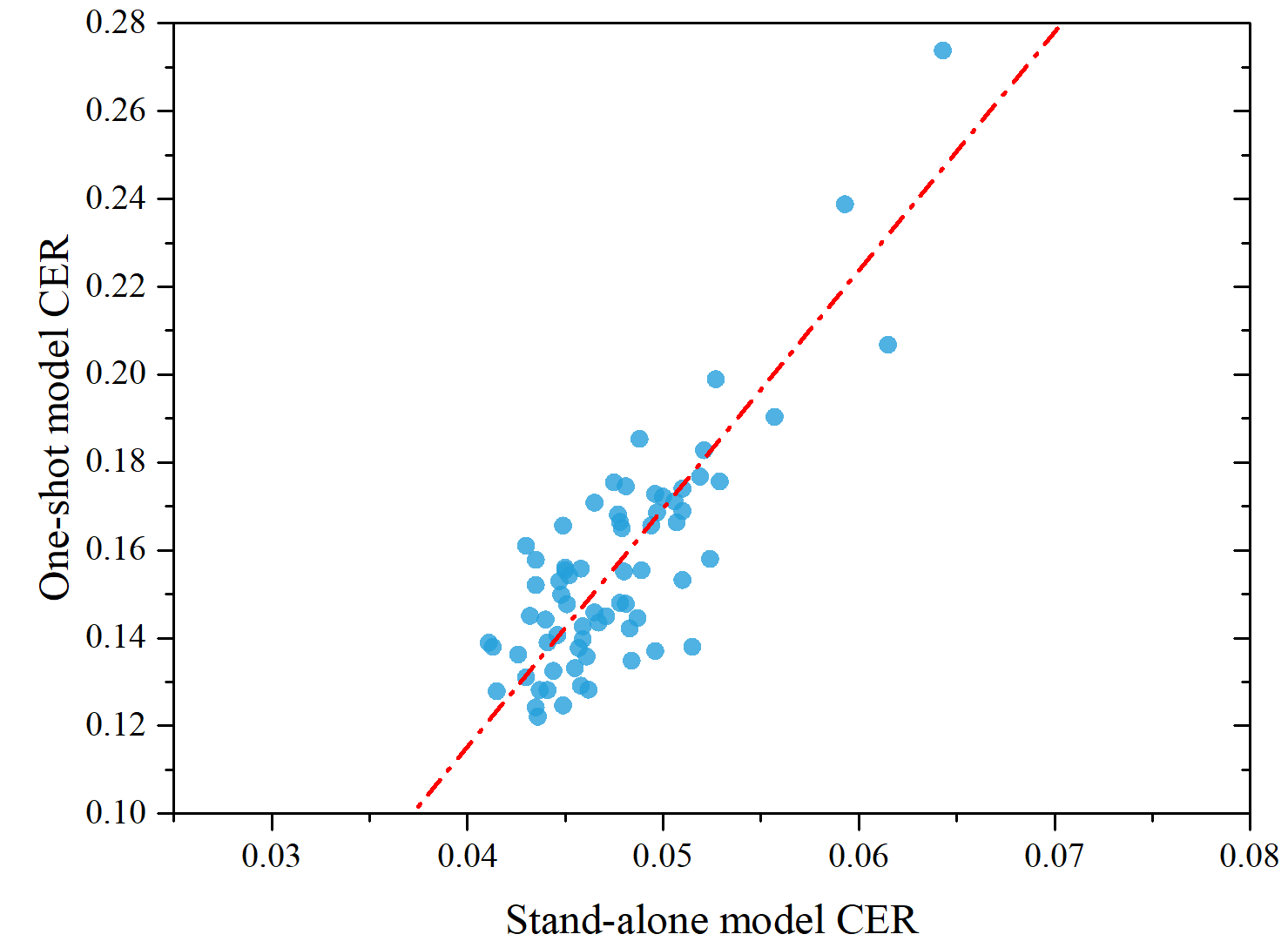}}
	\subfigure[Best path.]{\includegraphics[width=0.24\textwidth]{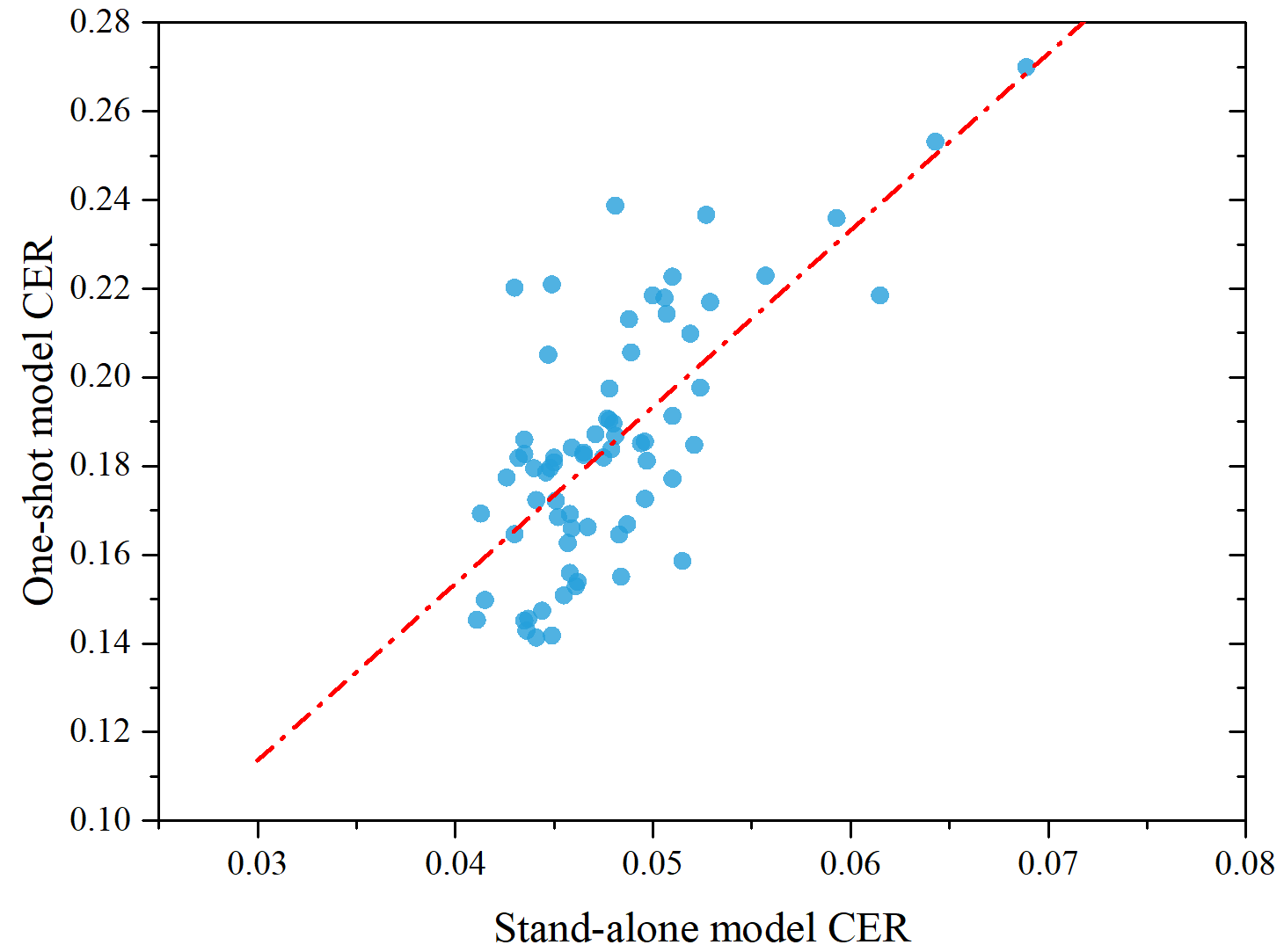}}
	\subfigure[Co-update.]{\includegraphics[width=0.24\textwidth]{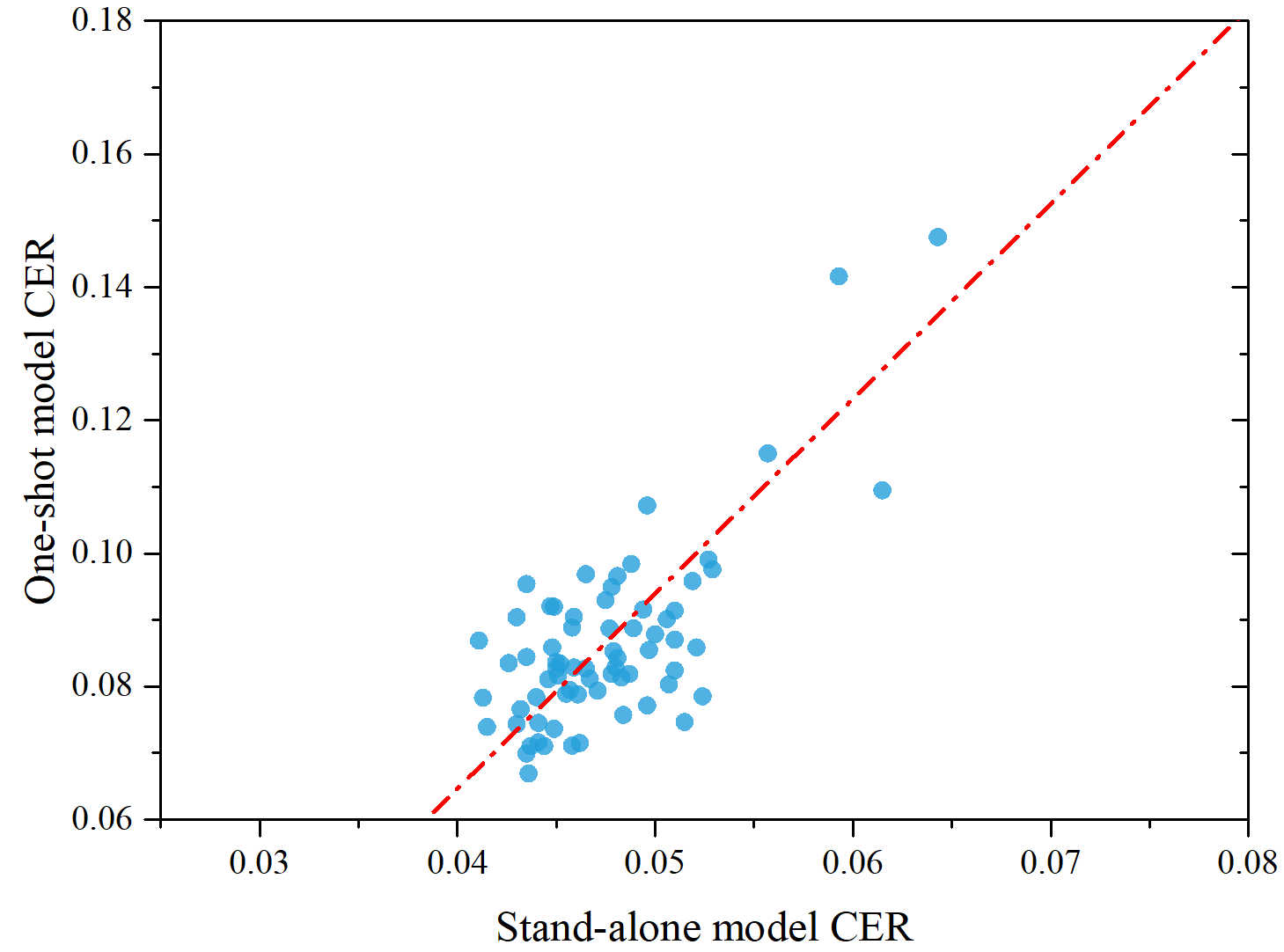}}
	
	\vspace{-10px}
	\caption{Correlation plots between the validation CER's of the stand-alone model and one-shot model for SPOS~\cite{guo2020single}, 
		TREFE (``random path") and its variants (``best path" and ``co-update"). 
		The red lines is the linear regression fit.}
	\label{fig:supernet_training2}
\end{figure*}

Figure~\ref{fig:supernet_training2} shows
the correlation plots for the four strategies, and
Table~\ref{tab:rank_supernet} shows
the ranking correlations.
As can be seen, the 
correlations are the smallest for 
SPOS, and highest for
``random path'',
demonstrating the advantage of the proposed strategy for sampling and updating $\bm{\alpha}_k^{\leftarrow}$.

\begin{table}[ht]
	\caption{Ranking correlations for different strategies.}
	\centering
	\vspace{-10px}
	\setlength\tabcolsep{4pt}
	\begin{tabular}{c|c|c|c}
		\toprule
		            & Kendall's $\tau$ & Spearman's $\rho$ & Pearson's $r$  \\ \midrule
		   SPOS     &      0.143       &       0.231       &     0.185      \\
		Random path &      0.501       &       0.686       &     0.871      \\
		 Best path  &      0.447       &       0.598       &     0.708      \\
		 Co-update  &      0.371       &       0.515       &     0.806      \\ \bottomrule
	\end{tabular}
	\label{tab:rank_supernet}
\end{table}

\begin{table*}[ht]
	\centering
	\caption{Mean and standard deviation of the performance of TREFE with five
	repetitions.}
	\vspace{-10px}
	\begin{tabular}{c|c|c || c|c|c || c|c}
		\toprule
\multicolumn{3}{c||}{IAM}            &           \multicolumn{3}{c||}{RIMES}          & \multicolumn{2}{c}{IIIT5K} \\
WER (\%)       &   CER (\%)      & latency (ms)       &   WER (\%)    &   CER
(\%)    & latency (ms)      & Acc (\%) &  latency (ms)  \\ \midrule
16.41$\pm$0.21 &   4.45$\pm$0.34 & 2.85$\pm$0.12  &   9.16$\pm$0.12    &     2.75$\pm$0.19      &     2.86$\pm$0.27      &   94.8$\pm$0.1   &  2.62$\pm$0.23      \\ \bottomrule
	\end{tabular}
	\label{reply_tab:variation}
	\vspace{-10px}
\end{table*}

\vspace{5px}
\noindent
\textbf{Effect of the Number of Blocks $K$.}
In this experiment, we perform ablation study on $K$. 
We train the supernet using
Algorithm~\ref{alg:prog} with $K=3,5,7$.
The ranking correlation between the validation set CER
of the
stand-alone 
and 
one-shot models 
is
shown in Table~\ref{tab:ablation_k}.
As can be seen, TREFE is robust to the value of $K$.
Besides,
the supernet training cost
decreases with $K$ (they are
2.0, 1.4, and 1.3 GPU days, 
for $K=3,5,7$,
respectively).
When $K$ increases, each block becomes smaller, and the training time reduction
in training smaller blocks is more significant than the larger number of blocks that
have to be trained.

\begin{table}[ht]
	\caption{Effect of $K$ on the ranking correlation.}
	\centering
	\vspace{-10px}
	\begin{tabular}{c|c|c|c}
		\toprule
$K$ & Kendall's $\tau$ & Spearman's $\rho$ & Pearson's $r$ \\ 
\midrule
		3           & 0.471   & 0.640  & 0.850    \\ \midrule
		5           & 0.501   & 0.686  & 0.871    \\ \midrule
		7           & 0.503   & 0.683  & 0.870    \\ 
		\bottomrule
	\end{tabular}
	\label{tab:ablation_k}
\end{table}


\subsubsection{Search on Supernet}
\label{sec:exp:abl_sea_alg}

\begin{figure}[ht]
	\centering
	\includegraphics[width=0.35\textwidth]{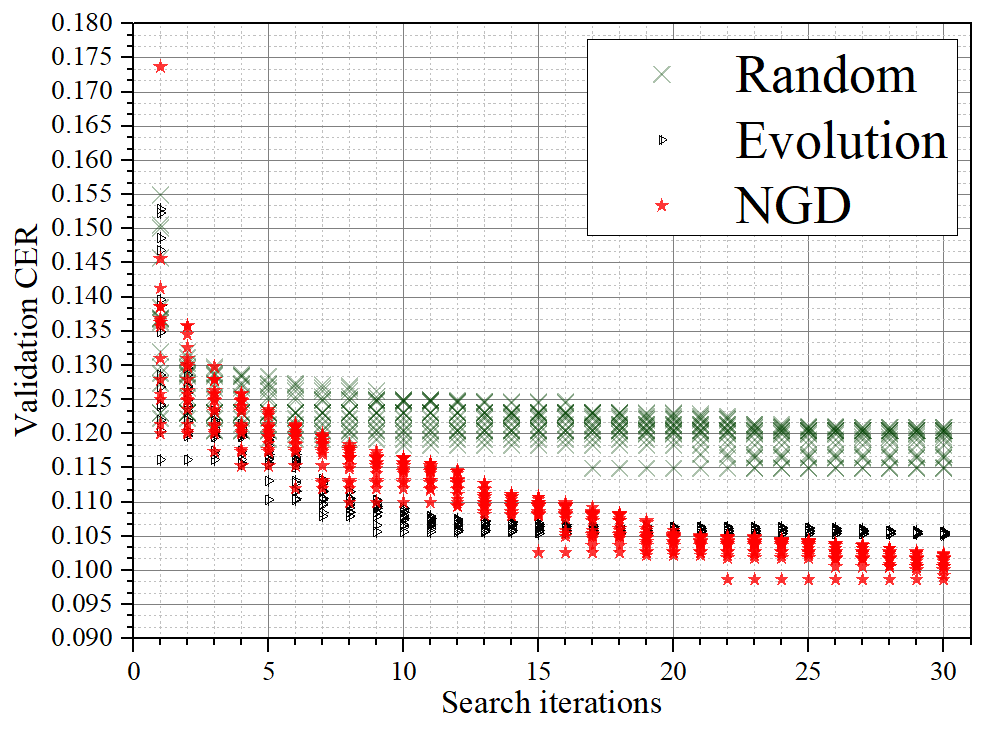}
	
	\vspace{-10px}
	\caption{Validation CER versus the number of search iterations for natural gradient descent (NGD),
	evolutionary algorithm and
	random sampling on the IAM dataset.}
	\label{fig:search_curve}
	\vspace{-10px}
\end{figure}

TREFE uses natural gradient descent  (denoted NGD)
to optimize the parameters 
$\bm{\theta}$ in \eqref{eq:search_on_supernet}. In this experiment, we compare
the efficiency of
NGD with 
random search and evolutionary  architecture search in~\cite{guo2020single}
on the IAM dataset.
In each search iteration, 16 new architectures are sampled.
Figure~\ref{fig:search_curve}
shows the 
number of search iterations versus the 
validation CERs
of the best 16 models obtained up to that iteration.
As can be seen,
NGD and evolutionary search clearly outperform random search,
and NGD performs the best.
In general, evolutionary algorithm can be easily trapped in local minima due to
inbreeding, and  so
the performance of
evolutionary search 
cannot improve after 10 search iterations.

As sampling is involved in the search,
i.e.,
\eqref{eq:fisher} and \eqref{eq:grad}, 
we study  the
variance of the proposed 
TREFE by running it
five times with different random seeds.
Table~\ref{reply_tab:variation} shows the mean and standard deviation of the
performance on the IAM, RIMES and IIIT5K datasets.
As can be seen, the
variance is small in all cases.

\subsection{Comparison with AutoSTR}
\label{sec:exp:comp}


Following the 
comparison between AutoSTR and TREFE
in Section~\ref{sec:diff:autostr},
in this section
we perform an ablation study on 
AutoSTR,
TREFE
and 
different variants.
Table~\ref{tab:autostr:iam} shows the comparison on handwritten text recognition using the IAM and RIMES datasets,
and 
Table~\ref{tab:autostr:iiit} shows the performance comparison on scene text recognition using the IIIT5K dataset.

\begin{itemize}[leftmargin=*]
\item 
On comparing AutoSTR with variant-1 in
Table~\ref{tab:autostr:iam}
(resp.
Table~\ref{tab:autostr:iiit}),
using sequential attention (SeqAtt) as the recognition head leads to much higher
latency and also higher error than
the use of
CTC
(resp. 
parallel attention)
in 
handwritten
(resp.  scene)
text recognition.

\item The improvement of variant-2 over variant-1 shows effectiveness of the searched transformer.

\item 
The improvement of
variant-3
over variant-1
shows 
the effectiveness of the proposed search algorithm 
for the spatial model.

\item Finally,
the improvement of TREFE over
variant-1  
shows the improvement with both the learned spatial and sequential models.
\end{itemize}

\begin{table*}[ht]
	\centering
	\caption{Comparison of 
	TREFE,
	AutoSTR
	and different variants on IAM and RIMES dataset.}
	\vspace{-10px}
	\begin{tabular}{c|c|c|c||c|c|c||c|c|c}
		\toprule
& spatial        &     sequential               &   recogition &
\multicolumn{3}{c||}{IAM} & \multicolumn{3}{c}{RIMES}          \\
& model & model & head & WER (\%) & CER (\%) & latency (ms) & WER (\%) & CER (\%)
&    latency (ms) \\ \midrule
TREFE		& $\mathcal{N}_{\text{spa}}^*$ & $\mathcal{N}_{\text{seq}}^*$ &    CTC    &  16.41   &   4.45   &     2.85     &   9.16   &   2.75   &     2.86     \\ \midrule
AutoSTR &         CNN         &            BiLSTM            &  SeqAtt   &  45.23   &  26.24   &    11.42     &  20.40   &  11.31   &    12.31     \\ \midrule
variant-1 &         CNN         &            BiLSTM            &    CTC    &  24.09   &   6.84   &     3.18     &  13.54   &   3.98   &     3.15     \\ \midrule
variant-2		&         CNN         & $\mathcal{N}_{\text{seq}}^*$ &    CTC    &  19.68   &   5.49   &     1.88     &  10.45   &   3.18   &     1.88     \\ \midrule
variant-3 & $\mathcal{N}_{\text{spa}}^*$ &            BiLSTM            &    CTC    &  18.65   &   5.01   &     4.03     &  10.47   &   3.20   &     4.04     \\ 
\bottomrule
	\end{tabular}
	\label{tab:autostr:iam}
	\vspace{-10px}
\end{table*}

\begin{table}[ht]
	\centering
	\caption{Comparison of 
	TREFE, AutoSTR and different
	variants on IIIT5K dataset.}
	\vspace{-10px}
	\setlength\tabcolsep{4pt}
	\begin{tabular}{c|c|c|c|c|c}
		\toprule
&           spatial            &          sequential          & recogition & Acc
& latency \\
&            model             &            model             &    head    & (\%) & (ms) \\ \midrule
TREFE       & $\mathcal{N}_{\text{spa}}^*$ & $\mathcal{N}_{\text{seq}}^*$ &   ParAtt   & 94.8 & 2.62 \\ \midrule
AutoSTR      &             CNN              &            BiLSTM            &   SeqAtt   & 94.7 & 3.86 \\ \midrule
variant-1 &             CNN              &            BiLSTM            &   ParAtt   & 93.3 & 2.01 \\ \midrule
variant-2 &             CNN              & $\mathcal{N}_{\text{seq}}^*$ &   ParAtt   & 93.6 & 1.53 \\ \midrule
variant-3  & $\mathcal{N}_{\text{spa}}^*$ &            BiLSTM            &   ParAtt   & 94.4 & 3.26 \\ \bottomrule
\end{tabular}
	\label{tab:autostr:iiit}
\end{table}


\section{Conclusion}

In this paper, 
we propose to find suitable feature extraction
for the text recognition (TR) by neural architecture search.
We first design a novel search space for the problem,
which fully explore the prior from such a domain.
Then,
we propose a new two-stages-based search algorithm,
which can efficiently search feature downsampling paths and 
convolution types for the spatial model,
and transformer layers for the sequential model.
Besides, our algorithm can be deployment aware,
and finding good architecture within specific latency constraints.
Experiments demonstrate that our searched models can greatly 
improve the capability of the TR pipeline 
and achieve state-of-the-art results on both 
handwritten and scene TR benchmarks. 

\section*{Acknowledgments}

This work was supported by the National Natural Science Foundation of China 62225603.

\bibliographystyle{IEEEtran}
\bibliography{bib}

\begin{IEEEbiography}[{\includegraphics[width = 1\textwidth]{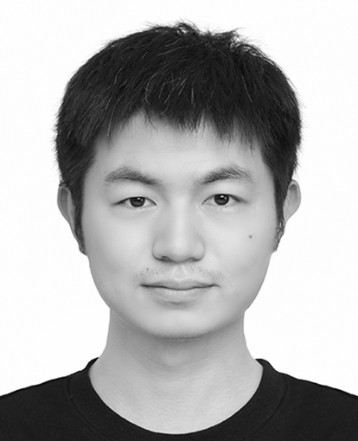}}]{Hui Zhang}
is currently a research engineer in  
4Paradigm Inc., Beijing, China. 
He received a B.S. degree 
in Computer Science and Technology
department 
from the HuaZhong University of Science and Technology (HUST), Wuhan, China, in 2018, 
and the 
M.S. degree in Electronics and Information Engineering 
also from HUST in 2020. 
His research interest is computer vision algorithms and systems.
\end{IEEEbiography}

\vspace{-10px}

\begin{IEEEbiography}[{\includegraphics[width = 1\textwidth]{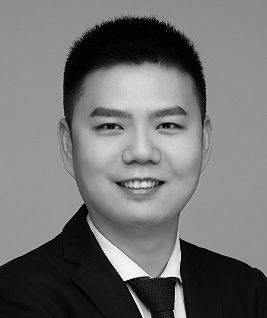}}]{Quanming Yao}
	(member, IEEE)
 is a tenure-track assistant professor in the Department of Electronic Engineering, Tsinghua University. 
He was a senior scientist in 4Paradigm, who is also the founding leader of the company's machine learning research team. He obtained his Ph.D. degree at the Department of Computer Science and Engineering of Hong Kong University of Science and Technology (HKUST). 
His research interests are in machine learning, graph neural networks, and automated machine learning. 
He is a receipt of Forbes 30 Under 30 (China), Young Scientist Awards (issued by Hong Kong Institution of Science), Wuwen Jun Prize for Excellence Youth of Artificial Intelligence (issued by CAAI), and a winner of Google Fellowship (in machine learning).
\end{IEEEbiography}

\vspace{-10px}

\begin{IEEEbiography}[{\includegraphics[width = 1\textwidth]{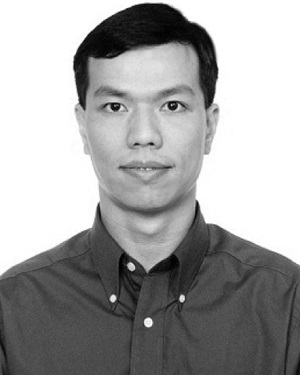}}]{James T. Kwok} (Fellow, IEEE)
	received the Ph.D. degree in computer science from The Hong Kong University of Science and Technology in 1996. 
	He is a Professor with the Department of Computer Science and Engineering, Hong Kong
	University of Science and Technology. His research interests include machine
	learning, deep learning, and artificial intelligence. He received the
	IEEE Outstanding 2004 Paper Award and the Second Class Award in Natural Sciences by
	the Ministry of Education, China, in 2008. 
	He is serving as an Associate Editor for the IEEE Transactions on Neural Networks and Learning Systems, Neural Networks, Neurocomputing, Artificial Intelligence Journal, International Journal of Data Science and Analytics, Editorial Board Member of Machine Learning,
	Board Member, and Vice President for Publications of the Asia Pacific Neural Network
	Society. He also served/is serving as Senior Area Chairs / Area Chairs of major machine learning /
	AI conferences including NIPS, ICML, ICLR, IJCAI, AAAI and ECML.
\end{IEEEbiography}

\vspace{-10px}

\begin{IEEEbiography}[{\includegraphics[width = 1\textwidth]{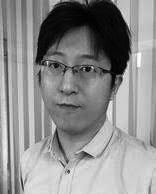}}]{Xiang Bai}
(Senior Member, IEEE)
received his B.S., M.S., and Ph.D.
degrees from the Huazhong University of Science and Technology (HUST), Wuhan, China,
in 2003, 2005, and 2009, respectively, all in
electronics and information engineering. He is
currently a Professor with the School of Artificial
Intelligence and Automation, HUST. He is also
the Vice-director of the National Center of AntiCounterfeiting Technology, HUST. His research
interests include object recognition, shape analysis, scene text recognition and intelligent systems. He serves as an associate editor for Pattern Recognition , Pattern
Recognition Letters, Neurocomputing and Frontiers of Computer Science.
\end{IEEEbiography}

\cleardoublepage
\appendices

\section{Obtaining the Number of Downsampling Paths}
\label{app:size}

Below Algorithm~\ref{alg:countpath},
which is used in Section~\ref{ssec:search_problem},
shows how to compute the number
of downsampling paths in the search 
space for the spatial model,
i.e.,
3D-mesh in Figure~\ref{fig:search_space},
through a recursive process (step 2-18).

\begin{algorithm}[H]
\caption{Backtracking algorithm to obtain the number of downsampling paths.}
\small
\begin{algorithmic}[1]
	\Require input size ($H_{\text{in}},W_{\text{in}}$), output size ($H_{\text{out}}, W_{\text{out}}$), stride $O_s$, number of layers $L$
	\State $n \gets 0$
	\Procedure{Backtracking}{$(h, w), l$}
		\If{$l = L$}
				\If{$h = H_{\text{out}} \; \textbf{and} \; w = W_{\text{out}}$}
					\State $n \gets n + 1$
				\EndIf
			\State \textbf{return}
		\EndIf

		\ForAll{$(s_h, s_w) \in O_s$}
			\State $h^\prime \gets h / s_h, w^\prime \gets w / s_w$
			\If{$h^\prime < H_{\text{out}} \; \textbf{or} \; w^\prime < W_{\text{out}}$}
				\State \textbf{continue}
			\EndIf
			\State $l^\prime \gets l + 1$
			\State \Call{Backtracking}{$(h^\prime,w^\prime), l^\prime$}
		\EndFor
		\State \textbf{return}
	\EndProcedure

	\State \Call{Backtracking}{$(H_{\text{in}},W_{\text{in}}), 0$}
	\State \textbf{return} $n$
\end{algorithmic}
\label{alg:countpath}
\end{algorithm}

\section{Training the Supernet with the Best Paths}
\label{app:best_path}
In the designed Algorithm~\ref{alg:prog} at Section~\ref{ssec:train_supernet}
for progressive training the big supernet,
$\bm{\alpha}_k$ (resp. $\bm{\alpha}_{k}^{\leftarrow}$) is uniformly sampled from 
$\Phi_k$ (resp. $\mathbb{S}_k$).
However,
one can also sample
the path which has best validation 
performance in $\mathbb{S}_k$ as $\bm{\alpha}_{k}^{\leftarrow}$ 
, as shown in step 5 of Algorithm~\ref{alg:train_supernet}.
To speed up the search for $\bm{\alpha}_{k}^{\leftarrow}$,
we construct a lookup table $\mathbf{Q}_k$ for each supernet block $\Phi_k$,
which records a pair of optimal structure and the corresponding performance, 
i.e., $(\bm{\alpha}^*, \text{perf}^*)$,
at different output resolutions $c$.
The procedure for searching $\bm{\alpha}_{k}^{\leftarrow}$ from $\mathbf{Q}$ 
is delineated in Algorithm~\ref{alg:best_path},
which greedy searchs $\bm{\alpha}_{k}^{\leftarrow}$ at block $j$ 
from $k-1$ to $1$ (step 4-17).
As the search space for $\Phi_k$ is small, we
can easily obtain $\mathbf{Q}_k$ using the trained supernet $\Phi_k$ (step 10-26).
The full procedure is in Algorithm~\ref{alg:train_supernet}
and compared as ``Best path'' in
Section~\ref{sec:exp:spos}.

\newpage

\begin{algorithm}[H]
	\caption{Train the supernet with best path.}
	\small
	\begin{algorithmic}[1]
		\Require $\Phi$, $K$, number of iterations $T$, Table $\mathbf{Q}$, maximum number of samples $E$
		\For{block $k$ = 1 \textbf{to} $K$}
			\State initialize output resolution lookup table $\mathbf{Q}_k \leftarrow \emptyset$
			\For{iteration $t$ = 1 \textbf{to} $T$}
				\State sample a path $\bm{\alpha}_k$ from $\Phi_k$ 
				\State $\bm{\alpha}_{k}^{\leftarrow} \gets$  \Call{BestPath}{$\mathbf{Q}, \bm{\alpha}_k$}
				\State sample a mini-batch $B_t$ from training data;
				\State update weights of $\bm{\alpha_k}$, the neck and the recognition head 
			\EndFor

			\State $e \gets 0$
			\For{$\bm{\alpha}_k$ in $\Phi_k$}
				\State $\bm{\alpha}_{k}^{\leftarrow} \gets$  \Call{BestPath}{$\mathbf{Q}, \bm{\alpha}_k$}
				\State $\text{perf} \gets \mathcal{A}_{\text{val}}(\mathbf{W}, (\bm{\alpha}_k^{\leftarrow}, \bm{\alpha}_k))$
				\State $c \leftarrow $ output resolution of $\bm{\alpha}_k$
				\If{$c$ not in $\mathbf{Q}_k$}
					\State $\mathbf{Q}_k[c] \gets (\bm{\alpha}_k, \text{perf})$
				\Else
					\State $(\bm{\alpha}_j^*, \text{perf}^*) \gets \mathbf{Q}_{j}[c]$
					\If{$\text{perf} > \text{perf}^*$} 
						\State $\mathbf{Q}_k[c] \gets (\bm{\alpha}_k, \text{perf})$
					
					\EndIf 
				\EndIf
				\State $e \gets e + 1$
				\If{$e > E$}
					\State \textbf{break} 
				\EndIf 
			\EndFor
		\EndFor	
	\State \textbf{return} $\mathbf{W}^*$ 
	\end{algorithmic}
	\label{alg:train_supernet}
\end{algorithm}

\begin{algorithm}[H]
	\caption{Finding a best path from $\mathbb{S}_k$.}
	\small
	\begin{algorithmic}[1]
		\Require  $\bm{\alpha}_k$, Table $\mathbf{Q}$
		\If{$k = 1$}
			\State \textbf{return} $\emptyset$
		\EndIf
		\For{block $j = K - 1$ \textbf{to} $1$}
			\State $r \gets$ input resolution of $\bm{\alpha}_{j+1}$
			\State $\bm{\alpha}_j^* \gets \emptyset, \text{perf}^* \gets -\inf $;
			\ForAll{stride $s$ in $O_s$}
				\State resolution $c$ goes from $r$ and inversed $s$
				\If{$c$ is invalid}
					\State \textbf{continue}
				\EndIf
				\State $(\bm{\alpha}_j, \text{perf}) \gets \mathbf{Q}_{j}[c]$
				\If{$\text{perf} > \text{perf}^*$}
					\State $\bm{\alpha}_k^{\leftarrow}[j] = \bm{\alpha}_j$
				\EndIf
			\EndFor
		\EndFor
		\State \textbf{return} $\bm{\alpha}_k^{\leftarrow}$
	\end{algorithmic}
	\label{alg:best_path}
\end{algorithm} 

\newpage

\section{Sampling Distribution $P_{\bm{\theta}}$}
\label{app:dist}

Consider a categorical variable $Y$ which takes values in 
$\{1,2,\dots,n_c\}$.
Consider the following probability distribution 
\begin{equation} \label{eq:tmp}
	p(Y=j;\theta)=\left\{
	\begin{array}{ll}
		\theta_j & 1 \leq j < n_c \\
		1 - \sum\nolimits_{i=1}^{j-1}\theta_i & j = n_c
	\end{array}
	\right.
\end{equation}
where
$\theta=[ \theta_1, \theta_2,\dots, \theta_{n_c-1}]$.
Let
$\mathbbm{1}\{\text{condition}\} \in \{0,1\}$ be the indicator function which
returns 1 when the condition holds, and 0 otherwise.
Then
(\ref{eq:tmp})
can be rewritten as
then 
\begin{align}
	p(Y=y;\theta) 
	&= \theta_1^{\mathbbm{1}\{y=1\}} \theta_2^{\mathbbm{1}\{y=2\}} \cdots \theta_{n_c}^{\mathbbm{1}\{y=n_c\}} \nonumber\\
	&= \theta_1^{(T(y))_1} \theta_2^{(T(y))_2} \cdots \theta_{n_c}^{1 - \sum\nolimits_{i=1}^{n_c - 1}(T(y))_i} \nonumber\\
	&= \exp(((T(y))_1)\log(\theta_1) + ((T(y))_2)\log(\theta_2) + \nonumber\\
	&\qquad \qquad \cdots + (1 - \sum\nolimits_{i=1}^{n_c - 1}(T(y))_i)\log(\theta_{n_c})) \nonumber\\
	&= \exp(\eta^\top T(y) - \phi(\eta)),
\end{align}
where 
$T(y)$ is the one-hot representation of $y$ without the last element,
$\eta=[\log(\theta_1/\theta_{n_c}), \cdots, \log(\theta_{n_c - 1}/\theta_{n_c})]$,
$\phi(\eta) = -\log(\theta_{n_c})$.

We model each choice of the search space 
(Section~\ref{sec:path_search_space} and Section~\ref{sec:rnn_search_space})
with the above probability distribution.
\begin{itemize}
	\item For the selection from convolution operations (i.e., $O_c$) and
	the selection from alternative choices in transformer layers (e.g., ``adding a residual path''),
	it is naturally set $n_c$ as 4 and 2 respectively.
	
	\item 
	For the selection from candidate stride (i.e., $O_s$),
	we simplify it.
	Specifically,
	to meet the downsampling path constrain in~\eqref{eq:size},
	stride $(2, 2)$ and $(2, 1)$ for downsampling 
	must appears 2 and 3 times respectively.
	Therefore, 
	we just need to determine the position $l \in [1, M]$ of those 5 strides.
	We set $n_c$ to be $M=20$ in the experiments,
	and discard some bad cases that 
	any two of the 5 strides have the same position.
\end{itemize}

\end{document}